\pdfoutput=1
\documentclass[10pt,journal,letterpaper,compsoc]{IEEEtran}

\usepackage[cmex10]{amsmath}
\usepackage{amssymb,bm,graphicx}
\usepackage{cite}
\usepackage[normalem]{ulem}
\usepackage[colorlinks=true]{hyperref}
\usepackage[T1]{fontenc}
\usepackage{rotating}
\usepackage{amsthm}
\usepackage{multirow}

\newtheorem{theorem}{Theorem}
\newtheorem{lemma}{Lemma}

\newtheorem{definition}{Definition}

\DeclareMathOperator*{\argmin}{arg\,min}

\usepackage{color}    

\newcommand{\ta}[0]{\widetilde{a}}

\usepackage{xspace}
\makeatletter
\DeclareRobustCommand\onedot{\futurelet\@let@token\@onedot}
\def\@onedot{\ifx\@let@token.\else.\null\fi\xspace}

\def\eg{\emph{e.g}\onedot} 
\def\ie{\emph{i.e}\onedot} 
 
\def\etc{\emph{etc}\onedot}

\makeatother

\begin{document}
\title{From Shading to Local Shape}

\author{Ying Xiong, Ayan Chakrabarti, Ronen Basri,\\Steven J. Gortler, David W. Jacobs, and Todd Zickler%
\IEEEcompsocitemizethanks{%
\IEEEcompsocthanksitem YX, AC, SJG, and TZ are with the Harvard School of Engineering and Applied Sciences, Cambridge, MA 02138, USA. RB is with the Weizmann Institute of Science, Rehovot 76100, Israel. DWJ is with the Dept. of Computer Science, University of Maryland, College Park, MD 20742, USA. E-mails: \{yxiong@seas.harvard.edu, ayanc@eecs.harvard.edu,  ronen.basri@weizmann.ac.il, sjg@cs.harvard.edu, djacobs@cs.umd.edu, zickler@seas.harvard.edu\}.}}

\IEEEcompsoctitleabstractindextext{
\begin{abstract}
We develop a framework for extracting a concise representation of the shape information available from diffuse shading in a small image patch. This produces a mid-level scene descriptor, comprised of local shape distributions that are inferred separately at every image patch across multiple scales. The framework is based on a quadratic representation of local shape that, in the absence of noise, has guarantees on recovering accurate local shape and lighting. And when noise is present, the inferred local shape distributions provide useful shape information without over-committing to any particular image explanation. These local shape distributions naturally encode the fact that some smooth diffuse regions are more informative than others, and they enable efficient and robust reconstruction of object-scale shape.  Experimental results show that this approach to surface reconstruction compares well against the state-of-art on both synthetic images and captured photographs.

\end{abstract}

}

\maketitle
\nocite{page}

\section{Introduction}
Recovering shape from diffuse shading is point-wise ambiguous because each surface normal can lie anywhere on a cone of directions. Surface normals are uniquely determined only where they align with the light direction which, at best, occurs at only a handful of singular points. A common strategy for reducing the ambiguity is to pursue global reconstructions of large, pre-segmented regions, with the hope that many point-wise ambiguities will collaboratively resolve, or that shape information will successfully propagate from identifiable singular points and occluding contours.

Global strategies are difficult to apply in natural scenes because diffuse shading is typically intermixed with other phenomena such as texture, gloss, shadows, translucency, and mesostructure. Occluding contours  and singular points are hard to detect in these scenes; and shading-based shape propagation breaks down unless occlusions, gloss, texture, \etc are somehow analyzed and removed by additional visual reasoning. Moreover, most global strategies do not provide spatial uncertainty information to accompany their output reconstructions, and this limits their use in providing feedback to improve top-down scene analysis, or in co-computing with other necessary bottom-up processes that perform complimentary analysis of other phenomena.

\begin{figure}[t]
\begin{center}
\includegraphics[width=\columnwidth]{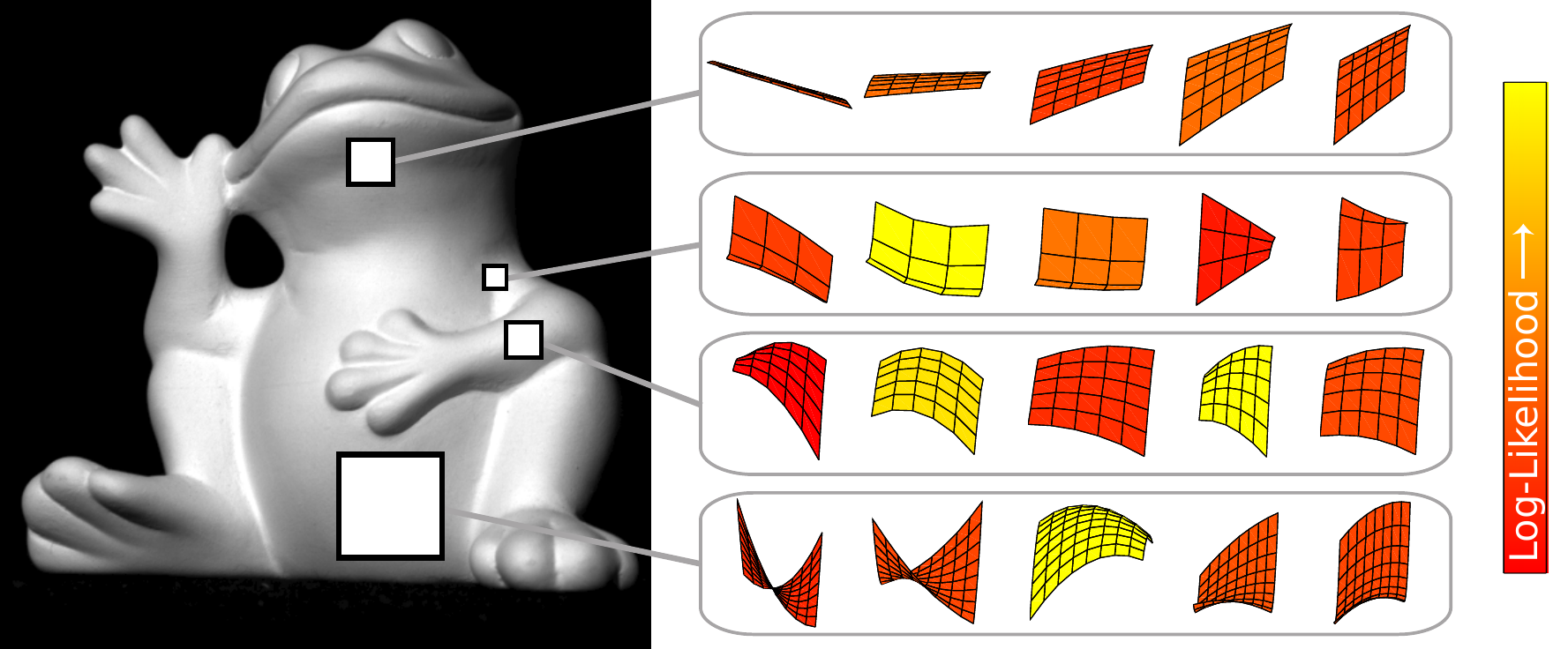}
\end{center}
\caption{\label{fig:teaser} We infer from a Lambertian image patch a concise representation for the distribution of quadratic surfaces that are likely to have produced it. These distributions naturally encode different amounts of shape information based on what is locally available in the patch, and can be unimodal (row 2 \& 4), multi-modal (row 3), or near-uniform (row 1). This inference is done across multiple scales.}

\end{figure}

This paper develops a framework for leveraging diffuse shading more broadly and
robustly by developing a richer description of what it says \emph{locally} about
shape. We show that point-wise ambiguity can be systematically reduced by
jointly analyzing intensities in small image patches, and that some of these
patches are inherently more informative than others. Accordingly, we develop an algorithm that produces for any image patch a concise distribution of surface patches that are likely to have created it. We propose these dense, local shape distributions as a new mid-level scene representation that provides useful local shape information without over-committing to any particular image explanation. Finally, we show how these local shape distributions can be combined to recover global object-scale shape.

Our framework is developed in three parts:

\setlength{\IEEEelabelindent}{0pt}
\setlength{\IEEEiednormlabelsep}{0.25em}
\setlength{\IEEEiedtopsep}{0pt}
\begin{enumerate}[{}]
\item\emph{Local uniqueness}. We provide uniqueness results for jointly recovering shape and lighting from a small image patch. By considering a world in which the shape of each small surface patch is exactly the graph of a quadratic function, we prove two generic facts: i) when the light direction is known, quadratic shape is uniquely determined; and ii) when the light is unknown, it is determined up to a four-way choice. We also catalog the degenerate cases, which correspond to special shapes, or conspiracies between the light and shape. These results are of direct interest to those studying the mathematics of shape from shading.
\item\emph{Local shape distributions}. We introduce a computational process that takes an image patch at any scale and produces a compact distribution of quadratic shapes that are likely to have produced it. At the core of this process is our observation that all likely shapes corresponding to a (noisy) image patch lie close to a one-dimensional manifold embedded in the five-dimensional space of quadratic shapes. This part of the paper is of broad interest because these local, multi-scale shape distributions may be useful as intermediate scene descriptors for various visual tasks.
\item\emph{Reconstruction}. We present a simple and effective bottom-up
  reconstruction system for inferring object-scale shape from a single image of
  a predominantly textureless and diffusely-shaded surface. This reconstruction
  system uses as input our local shape distributions inferred from dense,
  overlapping patches at multiple scales. It is conceptually simple,
  computationally parallelizable, and robust to non-idealities like shadows,
  texture, highlights, \etc, with reconstruction accuracy that compares well to
  the state of the art. This system is of direct interest to those studying
  algorithms for traditional shape from shading, and it is structured in a
  modular way that provides a step toward co-computation with other
  reconstructive processes that also analyze other phenomena.
\end{enumerate}

These three parts are tightly bound together. The uniqueness results in
Sec.~\ref{sec:noiseless_uniqueness} show that the quadratic model is a
particularly convenient representation for small surface patches. In the absence
of noise, both shape and lighting are locally revealed, and local shape is
generally unique when lighting is known.
Building on this, Sec.~\ref{sec:noise_instability} examines how
uniqueness breaks down in the presence of noise. While very different quadratic
shapes can produce equally-likely local intensity patterns, we find that all
highly-likely shapes lie close to a one-dimensional sub-manifold. Then,
Sec.~\ref{sec:stochastic} shows how to infer a dense set of sample shapes along
this sub-manifold, thereby taking an image patch and producing a one-dimensional
shape distribution. Finally, Sec.~\ref{sec:reconstruction} shows how these
multi-scale local distributions can enable robust global reconstruction of
shape, by naturally encoding the fact that some smooth diffuse regions are
more informative than others.

The project page associated with this paper~\cite{page} provides separate implementations of our algorithms for inferring local distributions (Sec.~\ref{sec:stochastic}) and global shape (Sec.~\ref{sec:reconstruction}). These are highly parallelized and can be executed on a single machine, a local cluster of machines, or a cluster from a standard utility computing service.

\section{Related Work}

Background on shape inference from diffuse shading can be found in several
reviews and serveys \cite{DurouFS:2008,HornB:1989,ZhangTCS:1999}. An important
question is whether shape is uniquely determined by a noiseless image, which has
been addressed by a variety of PDE-based formulations. For example, Oliensis
considered $C^2$ surfaces and showed that shape can be uniquely determined for
the entire image by singular points and occluding boundaries
together~\cite{Oliensis:1991}, and in many parts of the image by singular points
alone~\cite{Oliensis:1991b}. For the more general class of $C^{1}$ surfaces,
Prados and Faugeras~\cite{prados2005generic} employed a smoothness constraint to
prove uniqueness properties in a more general perspective
setup~\cite{PradosF:2004,PradosF:2005} given appropriate boundary conditions. In
this paper, we use a more restrictive local surface model but prove local
uniqueness without any boundary conditions or knowledge of singular points. This
generalizes previous studies of local uniqueness, which have considered
locally-spherical~\cite{pentland1984local} and
fronto-parallel~\cite{wagemans2010shading} surfaces.

Global uniqueness analyses have inspired global propagation and energy-based
methods for global shape inference
(\eg~\cite{DurouFS:2008,johnson2011shape,zhu2006shape}), some of which rely on
identifying occluding boundaries and/or singular points. While most methods do
not typically provide any measurement of uncertainty in their output, progress
toward representing shape ambiguity was made by Ecker and
Jepson~\cite{EckerJ:2010}, who use a polynomial formulation of global shape from
shading to numerically generate distinct global surfaces that are equally close
to an input image. In this paper, we study uniqueness and uncertainty at the
local level, and infer distributions over candidate local shapes.

Our work is related to patch-based approaches that use synthetically-generated
reference databases. The idea there is to reconstruct depth (or other scene
properties~\cite{freeman2000learning}) by synthesizing a database of aligned
image and depth-map pairs, and then finding and stitching together depth patches
from this database to match the input image and be spatially consistent. Hassner
and Basri~\cite{hassner2006example} obtain plausible results this way when the
input image and the database are of similar object categories, and Huang et
al.~\cite{huang2007examplar} pursue a similar goal for textureless objects using
a database of rendered Lambertian spheres. Cole et
al.~\cite{coles2012hapecollage} focus on patches located at detected keypoints
near an object's occlusion boundaries, combining shading and contour cues. We
also describe global shape as a mosaic of per-patch depth primitives, but
instead of relying on primitives from a pre-chosen set of 3D models, we consider
a continuous five-parameter family of depth primitives corresponding to graphs
of quadratic functions at multiple scales.

One of the our main motivations is the long-term goal of enabling better
co-computation with other bottom-up and top-down visual processes, and by
providing useful local shape information without choosing any single image
interpretation, our distributions are consistent with Marr's principle of least
commitment~\cite{marr1976early}. We focus on diffuse shading on textureless
surfaces, leaving for future work the task of merging with bottom-up processes
for other cues like occluding contours
(\eg,~\cite{coles2012hapecollage,huggins2001finding}), texture, gloss, and so
on. Our belief that this will be useful is bolstered by promising results
achieved by recent global approaches to such combined
reasoning~\cite{BarronM:2013}.

In independent work, Kunsberg and Zucker~\cite{kunsberg2013characterizing} have recently derived local uniqueness results that are related to, and consistent with, our results in Sec.~\ref{sec:noiseless_uniqueness}. Their elegant analysis, which uses differential geometry and applies to continuous images, is complimentary to the discrete and algebraic approach employed in this paper.
Kunsberg and Zucker also observe that the analysis of shading in patches instead of at isolated points is consistent with early processing in the visual cortex, and they discuss the possibility of local shading distributions being computed there. Indeed, the notion of such distributions is compatible with evidence that humans perceive shape in some diffuse regions more accurately than others~\cite{wagemans2010shading}.

\section{Quadratic-Patch Shape from Shading}
\label{sec:quad-patch-sfs}\label{sec:noiseless_uniqueness}

\newcommand{\xvec}{{\bar{x}}}
\renewcommand{\IEEEQED}{\IEEEQEDclosed}

We begin by analyzing the ability to uniquely determine the shape and lighting
of a local patch from a Lambertian shading image in the absence of noise. The
key assumption in our analysis is that depth of the patch can be \emph{exactly}
expressed as the graph of a quadratic function.
While subsequent sections consider deviations from this idealized setting, the
following analysis characterizes the inherent ambiguity under a local quadratic
patch model.

We model the depth $z(x,y)$ of a local surface patch as a quadratic function defined by coefficient vector $a\in \mathbb{Re}^5$ up to a constant offset:\footnote{Local shading for the special case $a_{4}=a_{5}=0$ is described in~\cite{wagemans2010shading}, and a more restrictive, locally-spherical model $z(x,y)=\sqrt{r^2-x^2-y^2}$ is analyzed in~\cite{pentland1984local}.}
\begin{equation}
\label{eq:zdefa}
z(x,y;a)=a_{1}x^{2}+a_{2}y^{2}+a_{3}xy+a_{4}x+a_{5}y.
\end{equation}
In matrix form, this is $z=[x,y] H [x,y]^T + J [x,y]^T$ with
\begin{equation}
\label{eq:hesdef}
H=\left[\begin{array}{cc}a_1 & a_3/2\\ a_3/2 & a_2\\ \end{array}\right]
\end{equation}
the Hessian matrix and $J=[a_4,a_5]$ the Jacobian of the depth function. The un-normalized surface normal to this patch at each location $(x,y)$ is then given by
\begin{equation}
\label{eq:ndef}
n(x,y;a)= [n_x(x,y;a),n_y(x,y;a),1]^{T},
\end{equation}
where
\begin{eqnarray}
  n_x(x,y;a) &\triangleq& -\frac{\partial z}{\partial x} = -2a_1x - a_3y - a_4,\label{eq:nxy-ax}\\
  n_y(x,y;a) &\triangleq& -\frac{\partial z}{\partial y} = -2a_2y - a_3x - a_5.  \label{eq:nxy-ay}
\end{eqnarray}
In matrix form, this is $n(x,y;a)=A[x,y,1]^T$ with
\begin{equation}
  \label{eq:Axdef}
  A\triangleq\left[\begin{array}{ccc}
      -2a_{1} & -a_{3} & -a_{4}\\
      -a_{3} & -2a_{2} & -a_{5}\\
      0 & 0 & 1
    \end{array}\right]
\end{equation}
the shape matrix corresponding to quadratic shape $a$.

The intensity $I(x,y;a)$ of this patch, observed from viewing direction
$v=[0,0,1]^T$ under a directional light source
$l=[l_{x},l_{y},l_{z}]^{T}$, is
\begin{equation}
\label{eq:measurement}
I(x,y;a)=\frac{l^{T}n(x,y;a)}{\|n(x,y;a)\|},
\end{equation}
assuming spatially-uniform Lambertian reflectance and that no part of the patch is in shadow, \ie, $l^T{}n(x,y) > 0, \forall
(x,y)$. Here, the magnitude $||l||$ of the light vector represents the product of the surface albedo and the light strength, and it is not assumed to be equal to one. Re-arranging,
the intensity $I$ at each point $(x,y)$ induces a quadratic constraint on its
surface normal~\cite{EckerJ:2010}:
\begin{equation}
\label{eq:measurement-quad}
I^2n^{T}{}n = n^{T}{}ll^{T}{}n \quad\Rightarrow\quad n^{T}\left(ll^{T}-I^2\mathbb{I}_{3\times 3}\right)n = 0,
\end{equation}
where $\mathbb{I}_{3\times 3}$ is the identity matrix.
This further induces a related constraint on shape parameters $a$:
\begin{equation}
  \label{eq:quad-a}
  \left[a^{T}~~~1\right]~\Big(D^{T}\left(ll^{T}-I^2\mathbb{I}_{3\times 3} \right)D\Big)~\left[ \begin{array}{c}a\\1\end{array}\right] = 0,
\end{equation}
where we use the matrix $D\in\mathbb{R}^{3\times 6}$ to re-write the relationship between $n$ and $a$ in \eqref{eq:ndef}-\eqref{eq:nxy-ay} as $n = D[a^{T}~~1]^{T}$.

Every pixel $(x,y)$ in an image patch gives one such constraint on shape parameters $a$, and shape from shading for quadratic patches rests on solving this system of polynomial equations. Our immediate goal is to determine whether the shape $a$ and lighting $l$ can be uniquely determined from these local constraints.

\subsection{Uniqueness of simultaneous shape and light}
\label{sec:generallight}

We assume that the local patch is sufficiently large to contain a minimum number
of \emph{non-degenerate} pixel locations, where the condition for non-degeneracy
is defined as follows:
\begin{definition}
For a patch $\Omega=\left\{ (x_{i},y_{i})\right\} _{i=1}^{N}$, we define the matrix $V_{\Omega} \in \mathbb{R}^{N\times 15}$ such that each row $v_i$ of $V_\Omega$ 
consists of all fourth-order and lower terms of $x_i$ and $y_i$:
\begin{equation}
\label{eq:qodef}
v_i = \left[x_i^4,~x_i^3y_i,~\ldots \underset{p,q \geq 0,~p+q\leq 4}{x_i^py_i^q} \ldots,x_i,~y_i,~1\right].
\end{equation}
A patch $\Omega$ is considered \emph{non-degenerate} if the matrix $V_\Omega$ has rank 15.
\end{definition}
Note that rectangular grids of pixels that are $5\times 5$ or larger will be non-degenerate under the above definition.

\begin{theorem}\label{prop:4-sol}
Given intensities $I(x,y)$ in an image patch $\Omega$ collected at a set of non-degenerate locations not in shadow, if any quadratic-patch/lighting pair $(a,l)$ that satisfies the set of polynomial equations \eqref{eq:quad-a} has a surface Hessian with eigenvalues that are not equal in magnitude, then there are no more than \textbf{four distinct surfaces} that can create the same image. Each of these surfaces is associated with a unique lighting when the Hessian of any solution is non-singular, and a one-dimensional family of lighting vectors otherwise.
\end{theorem}
This theorem states that given measurements of intensity from a quadratic surface patch, there generically exists four physical explanations, each comprised of a shape $a$, a light direction
$l/\|l\|$, and a scalar $\|l\|$ encoding the product of albedo and light
strength.  

Before proceeding to the proof, we introduce a lemma that relates to equations with ratios of quadratic terms. We define $\xvec \triangleq [x~~y~~1]^T$, so that the normals are given by $n(x,y;a) =  A\xvec$, and the intensity constraint \eqref{eq:quad-a} becomes
\begin{equation}
  \label{eq:isqdef}
  I_\xvec^2 = \left(\frac{l^Tn}{\|n\|} \right)^2 = \frac{\xvec^TA^Tll^TA\xvec}{{\xvec^TA^TA\xvec}}.
\end{equation}
Using this notation, we can state the following lemma, which is proven in the supplementary material:
\begin{lemma}
\label{lem:AllA}
Let $A$ and $\widetilde{A}$ correspond to two matrices of the form in \eqref{eq:Axdef}, and $l$ and $\widetilde{l}$ to two lighting vectors. If
\begin{equation}
  \label{eq:alla2}
  \frac{\xvec^TA^Tll^TA\xvec}{{\xvec^TA^TA\xvec}} = \frac{\xvec^T\widetilde{A}^T\widetilde{l}\widetilde{l}^T\widetilde{A}\xvec}{{\xvec^T\widetilde{A}^T\widetilde{A}\xvec}}, \forall \xvec \in \Omega,
\end{equation}
and if Rank$(V_\Omega) = 15$, Rank$(A) \geq 2$, and $l^TA\xvec> 0, \forall \xvec \in \Omega$ (i.e., no point is in shadow), then
\begin{equation}
\label{eq:alla}
A^Tll^TA = \widetilde{A}^T\widetilde{l}\widetilde{l}^T\widetilde{A},~~A^TA = \widetilde{A}^T\widetilde{A}.
\end{equation}
Moreover, if Rank$(A)=2$, then Rank$(\widetilde{A})=2$ and both $A$ and
$\widetilde{A}$ share a common null space.
\end{lemma}
\noindent {\bf Proof of Theorem \ref{prop:4-sol}}:~
Suppose there exists a solution $(a,l)$ that produces the observed set of intensities in the patch $\Omega$, and the Hessian matrix of surface $a$
has eigenvalues of un-equal magnitude. We will prove that if there exists another solution $(\widetilde{a},\widetilde{l})$, such that
\begin{equation}
\frac{\xvec^{T}A^{T}ll^{T}A\xvec}{\xvec^{T}A^{T}A\xvec}=I_{\xvec}^{2}=\frac{\xvec^{T}\widetilde{A}^{T}\widetilde{l}\ \widetilde{l}^{T}\widetilde{A}\xvec}{\xvec^{T}\widetilde{A}^{T}\widetilde{A}\xvec}, ~~\forall \xvec \in \Omega_i,\label{eq:quad-Al}
\end{equation}
then $\tilde{a}$ must be related to $a$ in one of four specific ways.

Since $a$ is not planar (otherwise the Hessian would have both eigenvalues equal
to zero), the corresponding matrix $A$ is at least rank 2, and we can apply
Lemma~\ref{lem:AllA}:
\begin{equation}
\widetilde{A}^{T}\widetilde{l}\ \widetilde{l}^{T}\widetilde{A}=A^T ll^T A,\qquad\widetilde{A}^{T}\widetilde{A}=A^{T}A.\label{eq:AllA}
\end{equation}

We define a new matrix $B$ satisfying $\widetilde{A} = BA$. Specifically, when $A$ is full rank we set $B=\widetilde{A}A^{-1}$; and when Rank$(A)=2$, we set $B=(\widetilde{A}+vv^T)(A+vv^T)^{-1}$ with $v$ a vector in the common null-space of $A$ and $\widetilde{A}$, \ie, $Av = \widetilde{A}v = 0$. We will show that there are only four possibilities for the matrix $B$.

Note that $A$ and $\widetilde{A}$ are affine matrices (last rows are both $[0,0,1]$). Moreover, in the rank 2 case, the last entry of $v$ will be 0 and $A+vv^T$ will also be an affine matrix. Therefore, $A^{-1}$ (if $A$ is full rank) and $(A+vv^T)^{-1}$ (if $A$ is rank 2) are affine. Hence, $B$ is also an affine matrix:
\begin{equation}
B=\left[\begin{array}{ccc}
b_{11} & b_{12} & b_{13}\\
b_{21} & b_{22} & b_{23}\\
0 & 0 & 1
\end{array}\right].\label{eq:B-affine}
\end{equation}
From (\ref{eq:AllA}), we have $B^{T}B=\mathbb{I}_{3\times3}$, \ie,
\begin{equation}
b_{13}^{2}+b_{23}^{2}+1=1\quad\Longrightarrow\quad b_{13}=b_{23}=0.
\end{equation}
The orthogonality of $B$ further restricts its top-left block
to be either a 2D rotation matrix
\begin{equation}
\label{eq:rotmat}
B = \left[\begin{array}{ccc}
\cos\varphi & -\sin\varphi & 0\\
\sin\varphi & \cos\varphi & 0\\
0 & 0 & 1
\end{array}\right],
\end{equation}
or an ``anti-rotation'' matrix
\begin{equation}
\label{eq:arotmat}
B = \left[\begin{array}{ccc}
\cos\varphi & \sin\varphi & 0\\
\sin\varphi & -\cos\varphi & 0\\
0 & 0 & 1
\end{array}\right],
\end{equation}
for $\varphi\in[-\pi,\pi)$.

From $\widetilde{A}=BA$ and the fact that the $(1,2)$-entry and
$(2,1)$-entry of $\widetilde{A}$ matrix should be the same (since
$a_{12}=a_{21}=-a_{3}$,$\widetilde{a}_{12}=\widetilde{a}_{21}=-\widetilde{a}_{3})$,
we have
\begin{equation}
2a_{1}b_{21}+a_{3}b_{22}=a_{3}b_{11}+2a_{2}b_{12}.\label{eq:integribility-constraint}
\end{equation}
This implies that when $B$ is of the form in \eqref{eq:rotmat}
\begin{equation}
  (a_1+a_2)\sin\varphi = 0,
\end{equation}
and when $B$ is of the form in \eqref{eq:arotmat}
\begin{equation}
  (a_1 - a_2) \sin\varphi = a_3 \cos\varphi.
\end{equation}
Since the Hessian of $a$ defined in \eqref{eq:hesdef} has eigenvalues of
un-equal magnitude, $a_1+a_2 \neq 0$, and either $a_1\neq a_2,$ or $a_3\neq
0$. This leaves only four possible solutions for $B$:
\begin{align}
\left[\begin{array}{ccc}
1 & 0 & 0\\
0 & 1 & 0\\
0 & 0 & 1
\end{array}\right],&
\left[\begin{array}{ccc}
\cos\varphi_{0} & \sin\varphi_{0} & 0\\
\sin\varphi_{0} & -\cos\varphi_{0} & 0\\
0 & 0 & 1
\end{array}\right],\notag\\
\left[\begin{array}{ccc}
-1 & 0 & 0\\
0 & -1 & 0\\
0 & 0 & 1
\end{array}\right],&
\left[\begin{array}{ccc}
-\cos\varphi_{0} & -\sin\varphi_{0} & 0\\
-\sin\varphi_{0} & \cos\varphi_{0} & 0\\
0 & 0 & 1
\end{array}\right],
\label{eq:B-4}
\end{align}
where $\varphi_{0}=\arctan\frac{a_{3}}{a_{1}-a_{2}}$. Thus $\widetilde{A}=BA$ can relate to $A$  in only four possible ways. 

Next, we consider the lighting $\widetilde{l}$ associated with each shape
$\widetilde{A}$. Equation (\ref{eq:AllA}) implies
$\widetilde{A}^T\widetilde{l}=A^Tl$ or $\widetilde{A}^T\widetilde{l}=-A^Tl$ but
the latter has shadows, so 
\begin{equation}\label{eq:ABl}
A^Tl = \widetilde{A}^T\widetilde{l} = A^TB^T\widetilde{l}.
\end{equation}
When $A$ is full rank, (\ref{eq:ABl}) implies a unique $\widetilde{l}$ given by
\begin{equation}
  \label{eq:ABL1}
  \widetilde{l} =(B^T)^{-1}l=Bl.
\end{equation}
If Rank$(A)=2$, we define $l_\perp$ as the component of $l$ in the null space of
$A^T$. Then, from (\ref{eq:ABl}), we have
\begin{equation}\label{eq:ABL2}
B^T\widetilde{l}=l+cl_\perp\quad\Rightarrow\quad\widetilde{l} = B(l+cl_\perp),
\end{equation}
where $c$ is a scalar. In this case there is a 1D family of $\widetilde{l}$ for each of the four shapes $\widetilde{A}$.
\hfill\IEEEQEDclosed

~

\begin{figure}[t]
\centering
\includegraphics[width=\columnwidth]{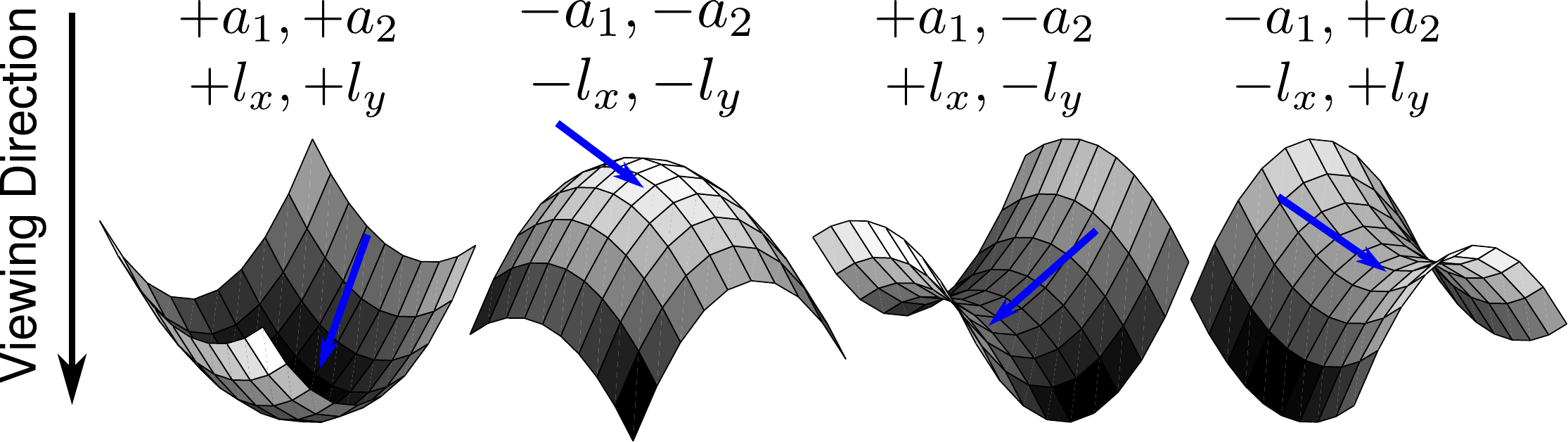}
\caption{Four quadratic-patch/lighting configurations that produce the same image (left is $a=[1,1/2,0,0,0],l=[2/3,1/3,2/3]$). The lighting is shown as blue arrows. The left pair and right pair are each convex-concave.\label{fig:canon-4-sol}}
\end{figure}

Figure~\ref{fig:canon-4-sol} provides an example of the four choices of shape/light pairs in the generic, non-cylindrical case when both eigenvalues of the surface Hessian are non-zero. Without loss of generality, we consider a rotated co-ordinate system where $a_3=0$, \ie, the $x$ and $y$ axes are aligned with the eigenvectors of the surface Hessian. Then, the four solutions from  \eqref{eq:B-4} are:
\begin{align}
\big([a_1,a_2,0,a_4,a_5],&~~[l_x,l_y,l_z]\big),\label{eq:choice1}\\
\big([-a_1,-a_2,0,-a_4,-a_5],&~~[-l_x,-l_y,l_z]\big),\label{eq:choice2}\\
\big([a_1,-a_2,0,a_4,-a_5],&~~[l_x,-l_y,l_z]\big),\label{eq:choice3}\\
\big([-a_1,a_2,0,-a_4,a_5],&~~[-l_x,l_y,l_z]\big).\label{eq:choice4}
\end{align}
The first choice is the surface/lighting pair $(a,l)$ that actually induced the image. The second corresponds to the well-known convex-concave ambiguity~\cite{pentland1984local}, and is obtained by reflecting both the light and the normals across the view direction. The last two choices \eqref{eq:choice3}-\eqref{eq:choice4} correspond to performing the reflection separately along each of the eigenvector directions of the Hessian matrix. These form a second concave-convex pair.

\begin{figure}[t]
\centering
\raisebox{1em}{\includegraphics[width=0.25\columnwidth]{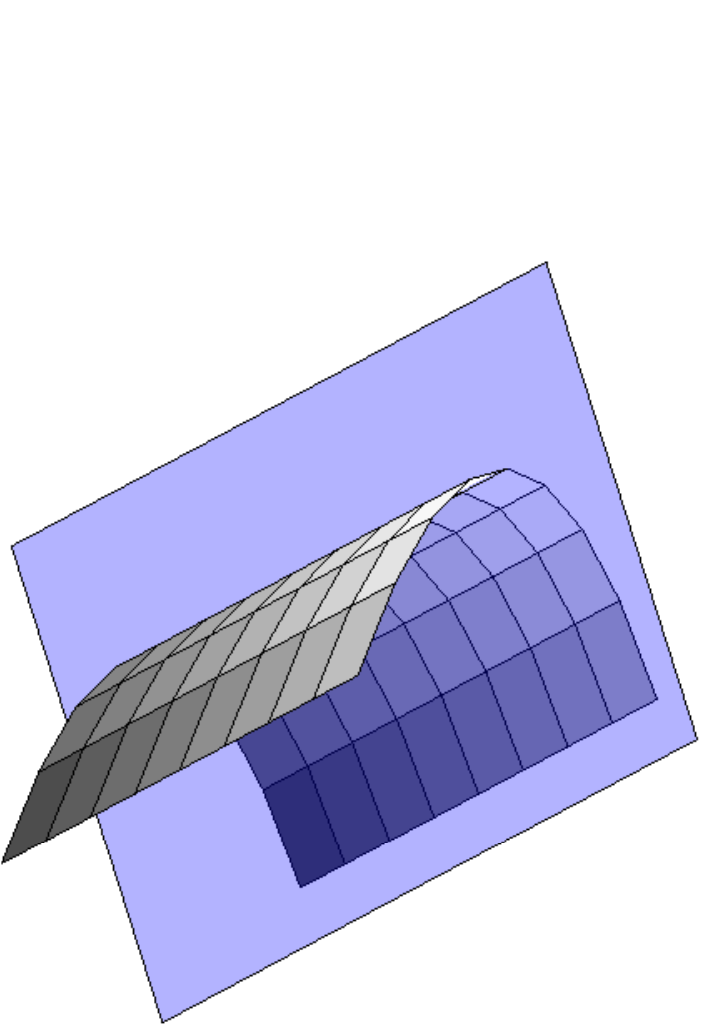}}
\includegraphics[width=0.20\columnwidth]{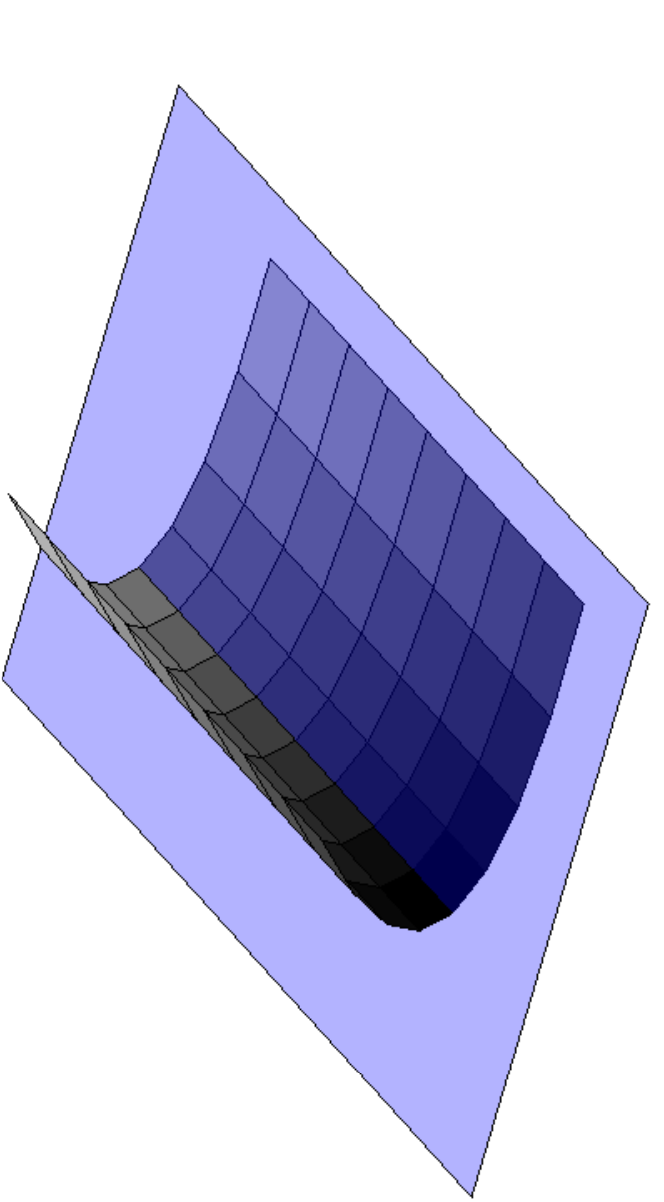}
\raisebox{1em}{\includegraphics[width=0.18\columnwidth]{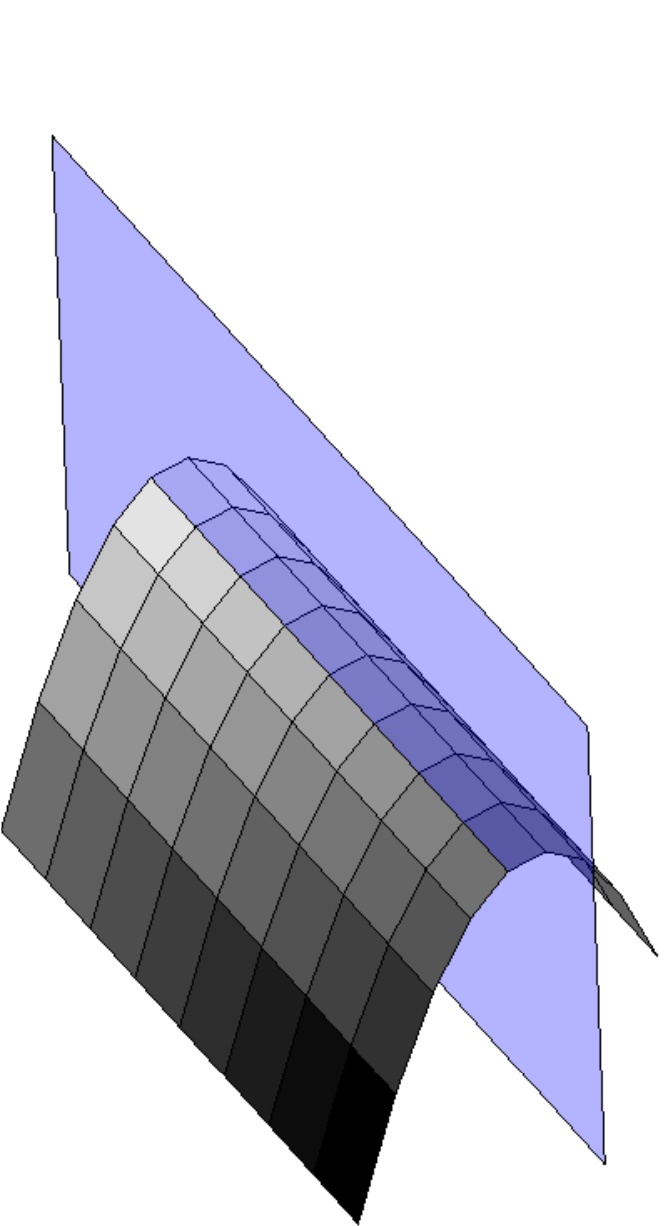}}
\raisebox{0.5em}{\includegraphics[width=0.22\columnwidth]{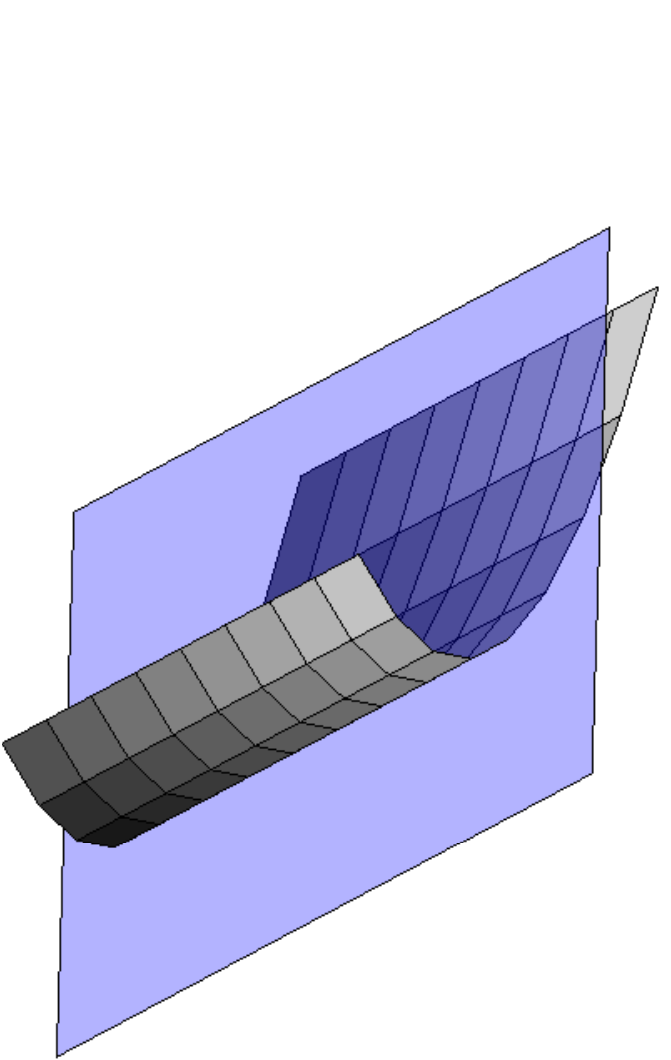}}
\caption{Lighting solutions in the cylinder case, when one of the eigenvalues of
  the surface Hessian is zero. There is a 1D family of lighting (any lighting
  direction in the blue plane with appropriate strength) for each of the four
  shapes that can produce the same image.}\label{fig:cylinder}
\end{figure}

When one of the Hessian eigenvalues is zero (say $a_2=0$ in our rotated
co-ordinate system), the patch surface is a cylinder and it is possible to
construct a 1D family of lights for each of the four surfaces:
\begin{equation}
\widetilde{l}=\textrm{diag}\{\textrm{sign}(\widetilde{a}_1a_1),
\textrm{sign}(\widetilde{a}_5a_5), 1\}\ (l + c\cdot[0, 1, a_5]^T)
\end{equation}
for any $c\in\mathbb{R}$ such that no pixel is in
shadow. Figure~\ref{fig:cylinder} shows an example of four cylindrical surfaces
and associated families of lights that can produce the same image.

Theorem \ref{prop:4-sol} applies when the Hessian eigenvalues of any solution
shape are not equal in magnitude. What happens when shape solutions have Hessian
eigenvalues that \emph{are} of equal magnitude? There are two distinct
cases. The first is when the Hessian is zero and the true surface is planar. In
this case every surface normal in the patch is identical, and the well-known
point-wise cone ambiguity applies to the patch as a whole: The observed image
can be explained by a one-parameter family of planar surfaces for \emph{every}
light $l$.

\begin{figure*}[t]
\centering
\includegraphics[width=\textwidth]{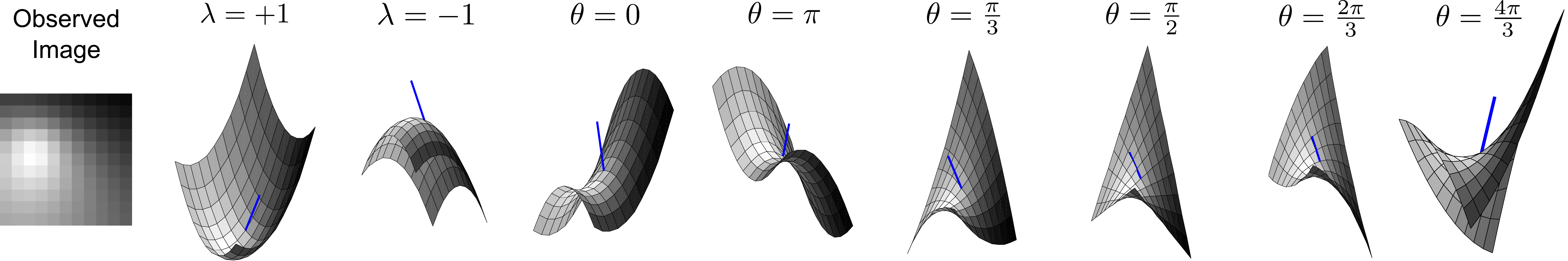}
\caption{When Hessian eigenvalues are equal in magnitude, there is a continuous family of patch/lighting pairs (given by \eqref{eq:hyperboloid-ambiguity-a} and \eqref{eq:hyperboloid-ambiguity-b}) that produce the same image. Note that the first four pairs above are analogous to Fig.~\ref{fig:canon-4-sol}.\label{fig:hyperboloid-ambiguity}}
\end{figure*}

In the second case, the true surface is not planar but the magnitudes of the two
eigenvalues of the Hessian matrix are equal. Unlike the planar ambiguity, there
is not an infinite number of surfaces that can combine with every lighting. But
as depicted in Fig.~\ref{fig:hyperboloid-ambiguity}, there is still an infinite
number of allowable patch/lighting pairs. We note that all quadratic surfaces in
this category can be expressed as either one of two following forms
\begin{align}
  \label{eq:hyperboloid-ambiguity-a}
a = & [r\cos \theta,-r\cos \theta,2r\sin \theta,\notag \\
    & \quad p\cos\theta-q\sin\theta,p\sin\theta+q\cos\theta],\\
\label{eq:hyperboloid-ambiguity-b}
a = & [\lambda r,\lambda r,0,\lambda p, -\lambda q],
\end{align}
where $\theta \in (-\pi,\pi], \lambda\in \{-1,+1\}, r \in \mathbb{R}^+$, and $p,q\in\mathbb{R}$. Given fixed values of $r,p$ and $q$, these surfaces generate identical images when paired with lighting
\begin{equation}
  \label{eq:hyperlight1}
  l=[l_{x}\cos\theta-l_{y}\sin\theta,l_{x}\sin\theta+l_{y}\cos\theta,l_z],
\end{equation}
for surfaces \eqref{eq:hyperboloid-ambiguity-a}, or with
\begin{equation}
  \label{eq:hyperlight2}
  l=[\lambda l_x,-\lambda l_y,l_z],
\end{equation}
for surfaces \eqref{eq:hyperboloid-ambiguity-b}, with fixed values of $l_x,l_y,l_z$.

\subsection{Unique shape when light is known}
\begin{theorem}
\label{cor:unique}
Given intensities $I(x,y)$ at a non-degenerate set of locations $\Omega$, a known light $l$, and a quadratic patch $a$ that satisfies the set of equations in \eqref{eq:quad-a}, if the planar component $[l_x,l_y]$ of the light is non-zero (i.e, $l$ is not equal to the viewing direction) and not an eigenvector of the Hessian of $a$, then the solution $a$ is unique.
\end{theorem}
\noindent {\bf Proof of Theorem \ref{cor:unique}}:~Without loss of generality, we choose a co-ordinate system where $a_3=0$. Note that for any such choice $l_x$ and $l_y$ will both be non-zero, unless $[l_x,l_y]$ is zero or an eigenvector of the surface Hessian, which is ruled out by the statement of the theorem. 

If the Hessian of $a$ has eigenvalues with unequal magnitudes, then it is easy to see that each of the four possible solutions from Theorem~\ref{prop:4-sol} has distinct light from \eqref{eq:ABL1} and \eqref{eq:ABL2}, and therefore for a fixed light, the shape is unique. A Hessian with \emph{equal} eigenvalues is ruled out since then every light-direction would be an eigenvector. When the eigenvalues have equal magnitudes but opposite signs, $a$ must be of the form in \eqref{eq:hyperboloid-ambiguity-a} with $\theta = 0$ or $\pi$ (since $a_3 = 0$) and $r = |a_1|=|a_2|$. In this case too, we see that each member of the continuous family of solutions---with $\theta \in (-\pi,\pi]$ for surface \eqref{eq:hyperboloid-ambiguity-a} and light \eqref{eq:hyperlight1}, or $\lambda \in \{-1,+1\}$ for surface \eqref{eq:hyperboloid-ambiguity-b} and light \eqref{eq:hyperlight2}---has a distinct light-direction. \hfill\IEEEQEDclosed

\begin{figure}[t]
\centering
\includegraphics[height=5.8em]{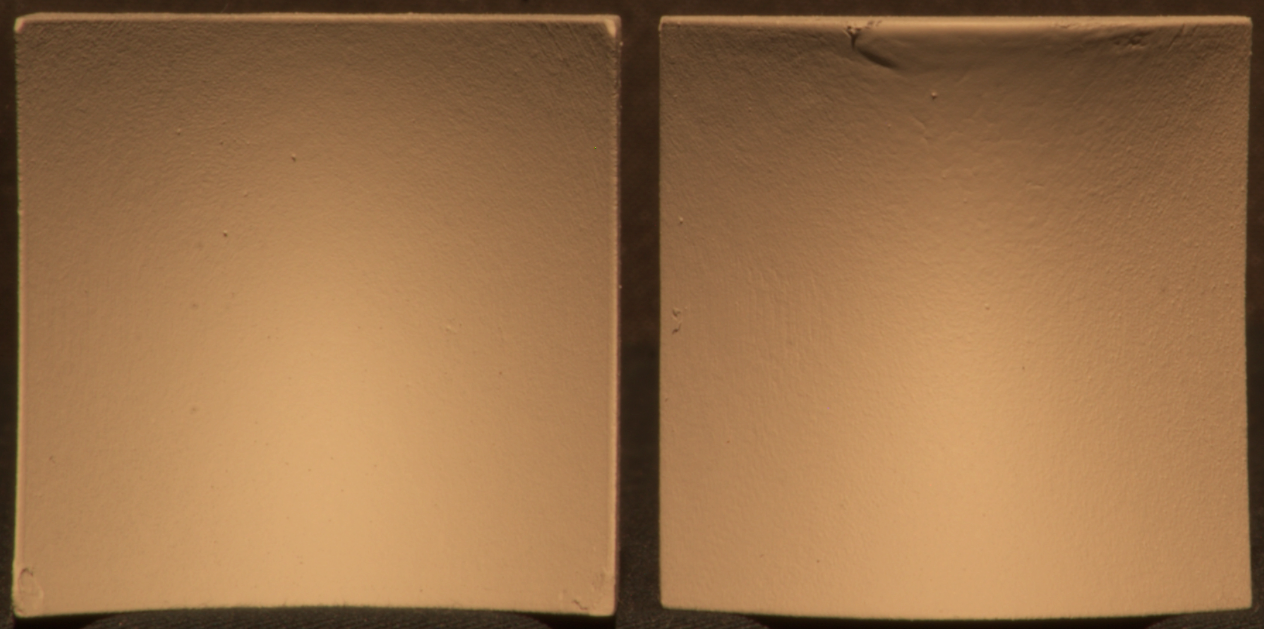}\hspace{2em}\includegraphics[height=5.8em]{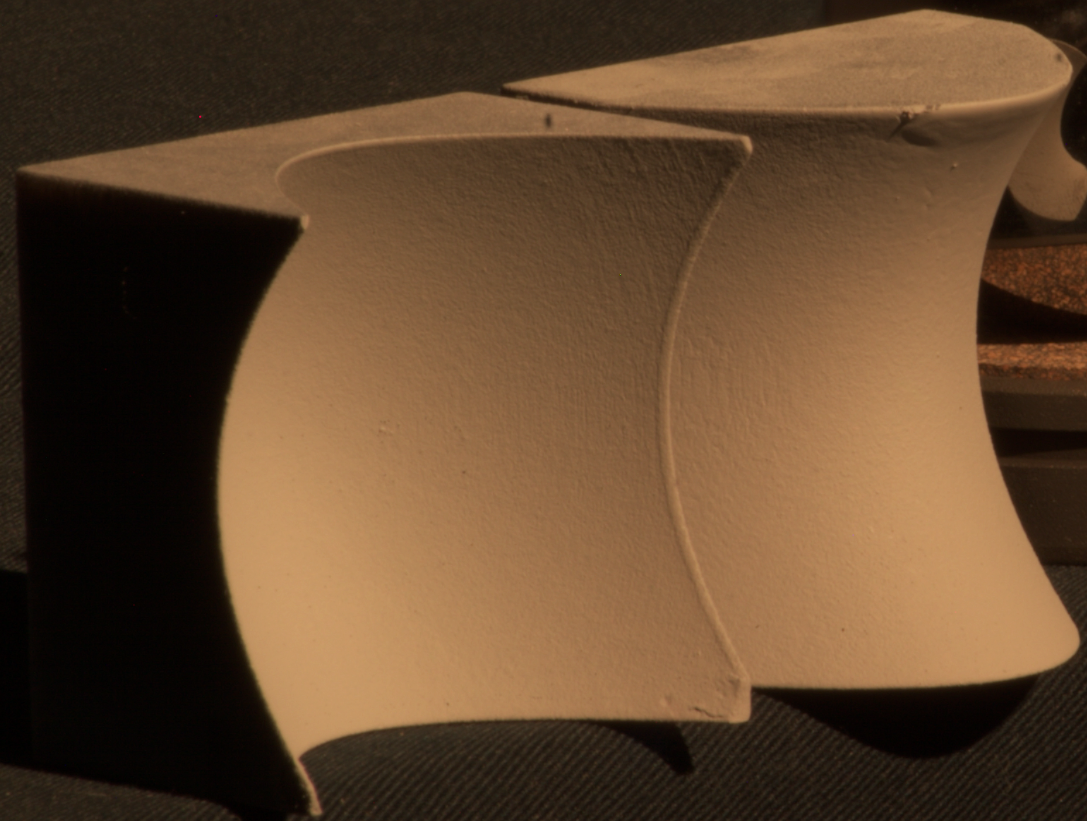}
\caption{Left: Two quadratic surfaces that produce the same image when the
  light is aligned with one of their common Hessian eigenvectors. For other view and light configurations (\eg, right) their images are distinct.
\label{fig:knownl-ambig}}
\end{figure}

When the conditions in Theorem~\ref{cor:unique} are not satisfied, there are shape ambiguities as follows. First, planar patches have Hessians with zero eigenvalues so that every $l$ is an eigenvector; this leads to an infinite set of planar shape explanations for any given light. Second, when the light and view directions are the same, there are generically four shape solutions analogous to Fig.~\ref{fig:canon-4-sol} or, in the case of equal eigenvalue magnitudes, a continuous family of solutions analogous to Fig.~\ref{fig:hyperboloid-ambiguity}. Finally, when the true surface is not planar but the azimuthal component of the light $[l_x,l_y]$ happens to be aligned with one of the Hessian eigenvectors, it is possible to construct a second solution by performing a reflection of the normals across that eigenvector direction. Figure~\ref{fig:knownl-ambig} demonstrates this with photographs of two 3D-printed surfaces that are distinct but related by a horizontal reflection of their normals.

\section{Ambiguity in the presence of noise}
\label{sec:noise_instability}\label{sec:local_info}

The uniqueness results from the previous section suggest that among the many possible models one could use for local shapes---such as splines, linear subspaces, exemplar dictionaries~\cite{huang2007examplar}, or continuous functions with smoothness constraints as in \cite{BarronM:2013}---the quadratic function model may be particularly useful. 
However, before we can use this model for inference, we must understand the effects of deviations, such as intensity noise and higher-order (non-quadratic) components of local shape. To this end, we provide some intuition about the types of quadratic shapes that \emph{almost} satisfy the polynomial system \eqref{eq:quad-a} and thus become likely explanations in the presence of noise. These intuitions motivate a statistical inference technique that will be introduced in Sec.~\ref{sec:stochastic}.

\begin{figure} 
\centering 
\includegraphics[width=1\columnwidth]{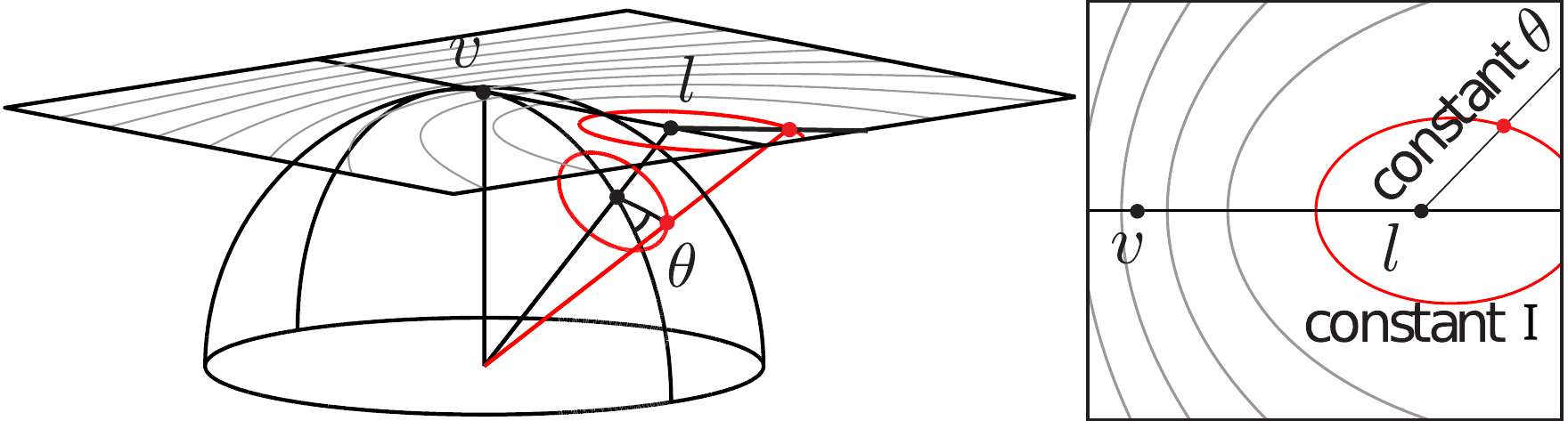} 
\caption{The light-centered cone of possible surface normals at any image point
  projects radially to a conic on the projective plane. We parameterize these
  conics by the radial projection of spherical angle $\theta$.
\label{fig:projective-plane}}
\end{figure}

In the rest of this paper, we assume that the light direction $l/\|l\|$ and the albedo/light-strength product $\|l\|$ are known.  Then, the polynomial system \eqref{eq:quad-a} relating the quadratic parameters $a$ to the observed intensities $I$ can be understood as combining two types of constraints on the patch normals $n=[n_x,n_y,1]$. First, each pixel's normal is constrained by its intensity to a light-centered circle of directions as per \eqref{eq:measurement}. This is shown in the left of Fig.~\ref{fig:projective-plane}, where the circle of directions is parameterized by ``azimuthal'' angle
\begin{equation}
\theta=\arctan\left(\frac{n_{x}l_{y}-n_{y}l_{x}}{l_{x}^{2}+l_{y}^{2}-l_{z}\left(n_{x}l_{x}+n_{y}l_{y}\right)}\right).\label{eq:theta(n)}
\end{equation}
 
The second type of constraint comes from the quadratic shape model, which induces a joint geometric constraint on the set of surface normals that belong to the patch. This joint  constraint has an intuitive interpretation when we represent the normals, light, and view as points on the plane defined by $n_z = 1$ (the so-called projective plane~\cite{TanQZ2011}). This representation is constructed  by radially-projecting the hemisphere of directions onto the plane as shown in Figure~\ref{fig:projective-plane}. The view is the origin of the plane, the light projects to another planar point $(l_x,l_y)/l_z$, and each pixel's $\theta$-parameterized circle of normal azimuthal directions projects to a conic section, still parameterized by $\theta$. The set of normals that lie on different conics but have the same azimuthal angle $\theta$ form a ray (right of Fig.~\ref{fig:projective-plane}), and an inversion in the sign of $\theta$ corresponds to a reflection of the surface normal across  light point.

\begin{figure}
\centering
\includegraphics[width=\columnwidth]{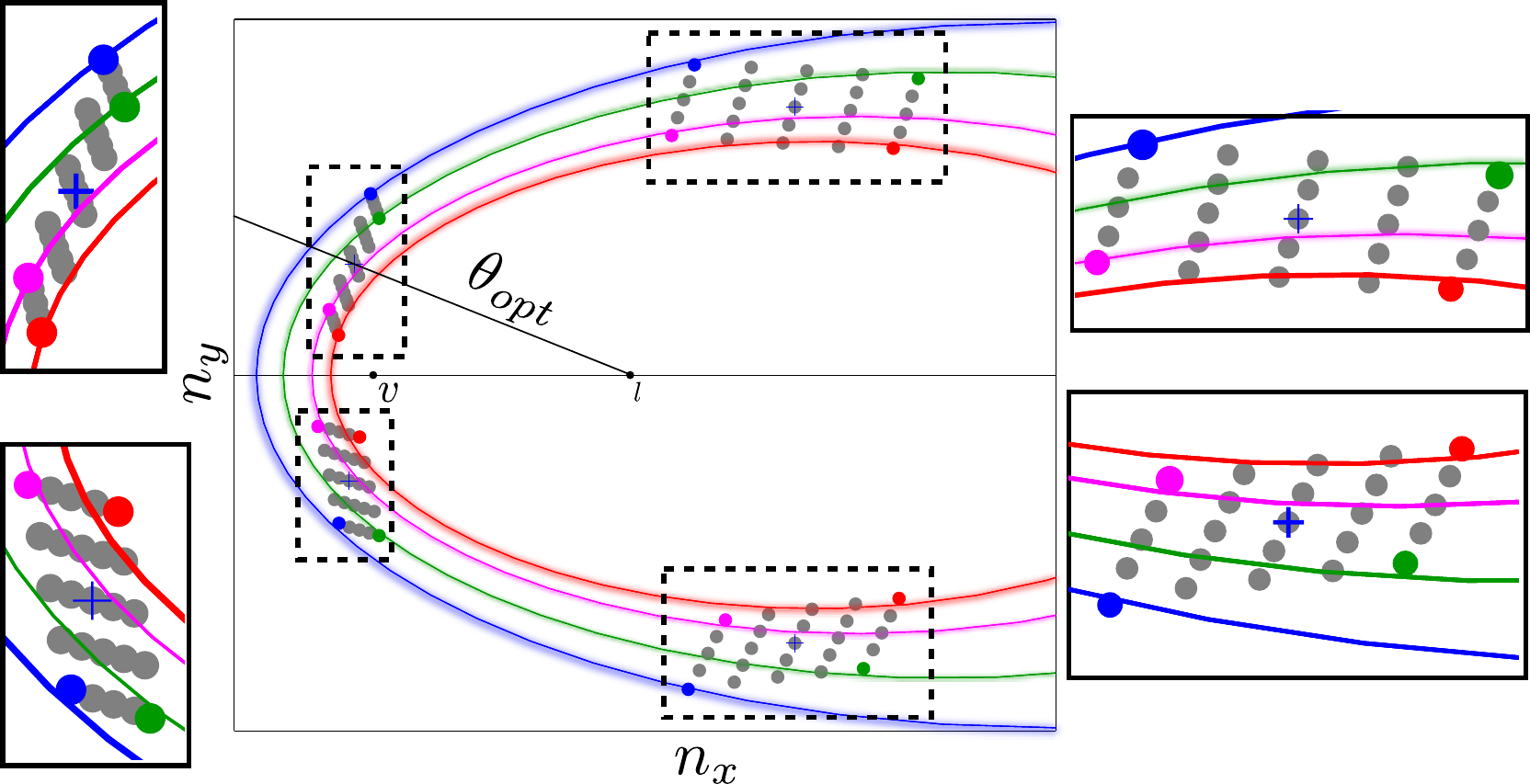}
\caption{Exact and approximate solutions for quadratic shape. Each color
  corresponds to a pixel in the patch (four are shown in the plot), whose
  intensity defines a conic curve that the normal vector should lie on. The
  normal vectors for a quadratic patch should form an affine grid on the
  projective plane, and good-fit shapes have grids that are well-aligned with
  the corresponding conics. The top left grid corresponds to an exact fit.
\label{fig:affine-grids}}
\end{figure}

Using this representation, Fig.~\ref{fig:affine-grids} visualizes the two types of constraints (under a light with $l_y=0$) for 25 normals at a $5\times 5$ grid of $(x,y)$ pixel locations. In addition to each pixel's normal being constrained to its conic, the set of normals is collectively constrained, via \eqref{eq:Axdef}, to be a symmetric affine grid. Therefore, solving the polynomial system for quadratic coefficients $a$ amounts to finding a symmetric affine grid that aligns properly with the per-pixel conics. Theorem~\ref{cor:unique} tells us there is only one grid that aligns perfectly, but as shown in the figure, there will be other grids that come close. When there is noise, the shapes corresponding to all of these grids become likely explanations, even though they are physically quite different from one another. To avoid over-committing, local inference systems must output distributions of shapes that encode this fact.

Then, a natural question is: do we need to search the entire five-dimensional
space of quadratic parameters $a$ to find all the likely approximate solutions?
To answer this question, we note that these approximate solutions are
intuitively expected to arise from the degenerate cases detailed in
Theorem~\ref{cor:unique}. For example, we find that these solutions often occur
in pairs corresponding to reflections across the light direction (\ie, across
the $x$ axis in Fig.~\ref{fig:affine-grids}), which would correspond to a second
exact solution if the light were a eigenvector of the surface Hessian. Remember
that the most ambiguous degeneracy is the one induced by the true surface being
planar, when all the conics overlap and there is a continuous set of solutions
whose normals can be parametrized by a single angle $\theta$ as per
\eqref{eq:theta(n)}.  Based on this intuition, we define $\theta(a)$ as the
first-order orientation of the shape $a$ to be the angle of the center normal,
and find empirically that it is sufficient to search along only a
one-dimensional manifold parametrized by this angle.

\begin{figure}
\centering
\includegraphics[width=\columnwidth]{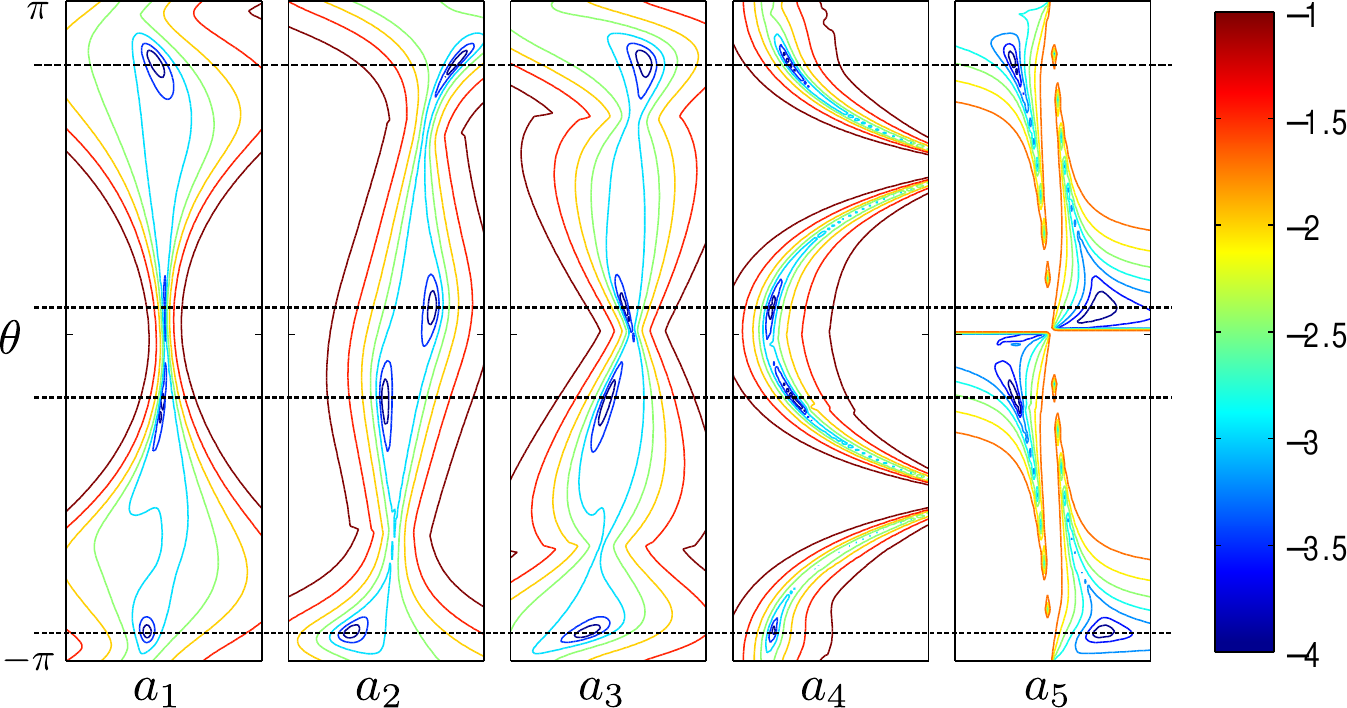}
\caption{Iso-contours of RMS intensity error for renderings of best-fit shape parameters $(a_1,a_2,a_3,a_4,a_5)$ when $\theta$ is fixed. Close fits occur at very different orientations (four modes here), but for any fixed orientation $\theta$ the remaining shape parameters are very constrained.
\label{fig:thvsAs}}
\end{figure}

In Fig.~\ref{fig:affine-grids}, this search can be understood as fixing the
value of $\theta(a)$, and warping an affine grid by optimizing the parameters
$a_1,a_2,a_3,a_4,a_5$ to fit the conic intensity constraints. We see that this
leaves very little play in the parameters, so the shapes $a$ of
possible solutions are highly constrained once $\theta(a)$ is fixed. This effect
is further visualized in Fig.~\ref{fig:thvsAs}, which shows contours of constant
RMS intensity difference---equally spaced in value on a logarithmic
scale---between the observed intensities and the Lambertian renderings of
best-fit shapes obtained by fixing $\theta(a)$ and one coefficient (say, $a_1$)
and then optimally fitting the others (say, $a_2,a_3,a_4,a_5$).  The four
``close fits'' appear as the four modes in these plots, where the value of
$\theta(a)$ strongly constrains each coefficient of low-error shapes $a$.

\section{Local shape distributions}
\label{sec:stochastic}

Armed with intuition about the characteristics of approximate solutions for the quadratic-patch model, we now develop a method for inferring shape distributions at any local image patch of any size. The output for each image patch is a set of quadratic shapes of the same size that correspond to a discrete sampling along a $\theta$-parametrized one-dimensional manifold, as well as a probability distribution over this set of quadratic shapes. The previous sections have demonstrated that shading in some image patches is inherently more informative than others. Our goal is to create a compact description of this ambiguity in each local region at multiple scales, thereby providing a useful mid-level representation of ``intrinsic'' scene information for vision.

\subsection{Computing quadratic shape proposals}\label{sec:shape-proposal}

Given the intensities $I_o(x,y)$ at a patch $(x,y) \in \Omega$, we first generate a set of quadratic proposals for the shape of that patch, and based on the intuition from the previous section, we index these proposals angularly in reference to the light $l$. Consider a discrete set of uniformly-spaced values $\theta^j,\ j \in \{1,\ldots J\}$ over $(-\pi,\pi]$\footnote{For some patches, we consider closer-spaced samples over a shorter interval when values close to $\pm \pi$ do not correspond to physically feasible estimates for shape.}%
, and for each angle $\theta^j$ we find the corresponding quadratic shape $a^j$ that best explains the observed intensities $I_o(x,y)$ in terms of minimum sum of squared errors:
\begin{equation}
  \label{eq:minerr}
  a^j = \argmin_{a: \theta(a) = \theta^j}~~\sum_{(x,y)\in\Omega} \|I_o(x,y) - I(x,y;a)\|^2,
\end{equation}
where $I(x,y;a)$ is defined as per \eqref{eq:measurement}.

Let $(0,0)$ be the center of the patch. Then since $\theta(a_i)$ is fixed, the quadratic coefficients $a_4$ and $a_5$ of $a^j$ only have one degree of freedom, and can be re-parametrized in terms of a single variable $r \in \mathbb{R}^+$ that indexes points along the constant $\theta$ ray on the projective plane:
\begin{align}
  a_4 &= -\frac{l_x}{l_z} - r\left(-\frac{l_x}{l_z}\cos\theta^j + l_y\sin\theta^j\right),\\
  a_5 &= -\frac{l_y}{l_z} - r\left(-\frac{l_y}{l_z}\cos\theta^j - l_x\sin\theta^j\right).
\end{align}
Therefore, the non-linear least-squares minimization in \eqref{eq:minerr} is
over the four variables $a_{1:3},r$, and can be efficiently carried out with
Levenberg-Marquardt~\cite{levenberg}. We found empirically that it is
insensitive to initialization, and use $[0,0,0,r_0]$ in our experiments, where
$r_0$ is chosen such that the center pixel lies on the corresponding conic.

This minimization occurs independently and in parallel for every patch in an image, and it can therefore be parallelized over an arbitrary number of CPU cores, on a single machine or a cluster of machines, as required for increasing image sizes. Our reference implementation considers $J=21$ quantized angles for each patch, and takes one minute on an eight-core machine for inference on all overlapping $5\times 5$ patches in a $128\times 128$ image.

\subsection{Computing shape likelihoods}

Next, we define a probability distribution over these shape proposals by
computing the likelihoods for the observed intensities being generated by each
proposed shape $a^j$. We introduce a model for the deviation between observed
intensity $I_o(x,y)$ and expected intensity $I(x,y;a)$ from a proposal $a$ at
each location $(x,y)$:
\begin{equation}
  \label{eq:pdfdef}
  I_o(x,y)\left| a \right. \sim \mathcal{N}\Big(I(x,y;a);~\sigma_i^2+\sigma_{z}^2(x,y;a)\Big).
\end{equation}
This is a Gaussian distribution conditioned on $a$, where the variance at each pixel $(x,y)$ is a sum of two terms. The first is additive \emph{i.i.d.}~intensity noise $\sigma_i^2$ induced, for example, by sensor noise. The second is a function of $a$ and varies spatially across the patch, capturing the fact that the veridical shape may exhibit higher-order (non-quadratic) variations at this patch's scale. It is the expected variance in intensity $\sigma_{z}^2(x,y;a)$ induced by higher-order components of shape that exist on top of the shape predicted by $a$ at the current scale.

\begin{figure*}[!t]
  \centering
  \includegraphics[width=\textwidth]{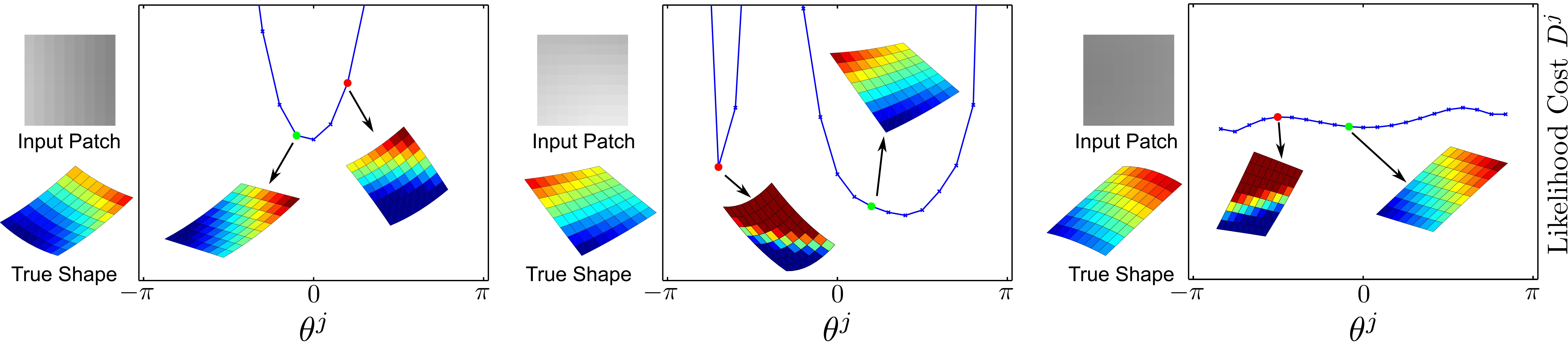}
  \caption{Shape likelihood distributions inferred from image patches. Graphs show likelihood cost $D^j$ over first-order orientation $\theta^j$, each of which is associated with a shape proposal $a^j$.}
  \label{fig:logpa}
\end{figure*}

To compute $\sigma_{z}^2(x,y;a)$, we model the deviations of the true normals $(\tilde{n}_x,\tilde{n}_y)$ from those predicted by $a$ as \emph{i.i.d.}~Gaussian random variables:
\begin{align}
  \tilde{n}_x(x,y) \sim \mathcal{N}(n_x(x,y;a); \sigma_{n0}^2),\\
  \tilde{n}_y(x,y) \sim \mathcal{N}(n_y(x,y;a); \sigma_{n0}^2),
\end{align}
where $\sigma_{n0}^2$ is the expected normal variance of these deviations, which is set to $10^{-6}$ in our experiment. Then, we compute $\sigma_{z}^2(x,y;a)$ as the expected variance in intensity over the distribution of $\tilde{n}_x,\tilde{n}_y$:
\begin{align}
  \label{eq:sigmasdef}
  &\sigma_{z}^2(x,y;a) = \underset{~~\tilde{n}_x,\tilde{n}_y}{\mathbb{E}}~~\left\|I(x,y;a) - \frac{l_x\tilde{n}_x+l_y\tilde{n}_y+l_z}{\sqrt{\tilde{n}_x^2+\tilde{n}_y^2+1}} \right\|^2\notag\\
& = \underset{~~\tilde{n}_x,\tilde{n}_y}{\mathbb{E}}~~ \left\|\frac{l_xn_x(a)+l_yn_y(a)+l_z}{\sqrt{n_x^2(a)+n_y^2(a)+1}} - \frac{l_x\tilde{n}_x+l_y\tilde{n}_y+l_z}{\sqrt{\tilde{n}_x^2+\tilde{n}_y^2+1}} \right\|^2.
\end{align}
We find that for lights not aligned with the view, \ie, $|l_z| < 1$, this expression can be reliably approximated as: 
\begin{equation}
\sigma_{z}^2(x,y;a) \approx \frac{(l_x^2+l_y^2)\sigma_{n0}^2}{n_x^2(x,y;a)+n_y^2(x,y;a)+1}.
\end{equation}
Intuitively, this says  that the observed intensity is less sensitive to perturbations in the normal $[n_x,n_y]$ when the surface is tilted away from the viewing direction.

Putting everything together, we compute a cost $D^j$ for every proposal $a^j$, defined as the negative log-likelihood of all observed intensities under the above model:
\begin{align}
  \label{eq:dcost}
  D^j=-\log p(I_o | a^j) =\sum_{(x,y)\in\Omega}& \frac{1}{2}\Big[~\log\left(\sigma_i^2+\sigma_{z}^2(x,y;a^j)\right)\notag\\
  &~+ \frac{\left(I_o(x,y)-I(x,y;a^j)\right)^2}{\sigma_i^2+\sigma_{z}^2(x,y;a^j)}~\Big].
\end{align}

\subsection{Evaluation}
\label{sec:local-evaluation}

We evaluate the accuracy of the proposed local shape distributions using images of a set of six random surfaces synthetically rendered (with the light at an elevation of $60^\circ$) , where each surface is created by generating a $5\times 5$ grid of random depth values, and then smoothly interpolating these to form a $128\times 128$ surface (see~\cite{page}, and Fig.~\ref{fig:SRex} for an example). 

Figure~\ref{fig:logpa} shows likelihood distributions and proposed shapes for
representative image patches from this synthetic dataset. Empirically, we find
that the distributions $D^j$ can have a single peak (left), be multi-modal
(center), or nearly uniform (right); re-enforcing our intuition that shading in
some image patches is more informative than others. Note that given the correct
value of $\theta^j$ (green dot in the figure), the corresponding estimated shape
proposal $a^j$ yields an accurate reconstruction of the true shape in all three
cases shown.

\begin{figure*}[!t]
  \centering
  \includegraphics[width=\textwidth]{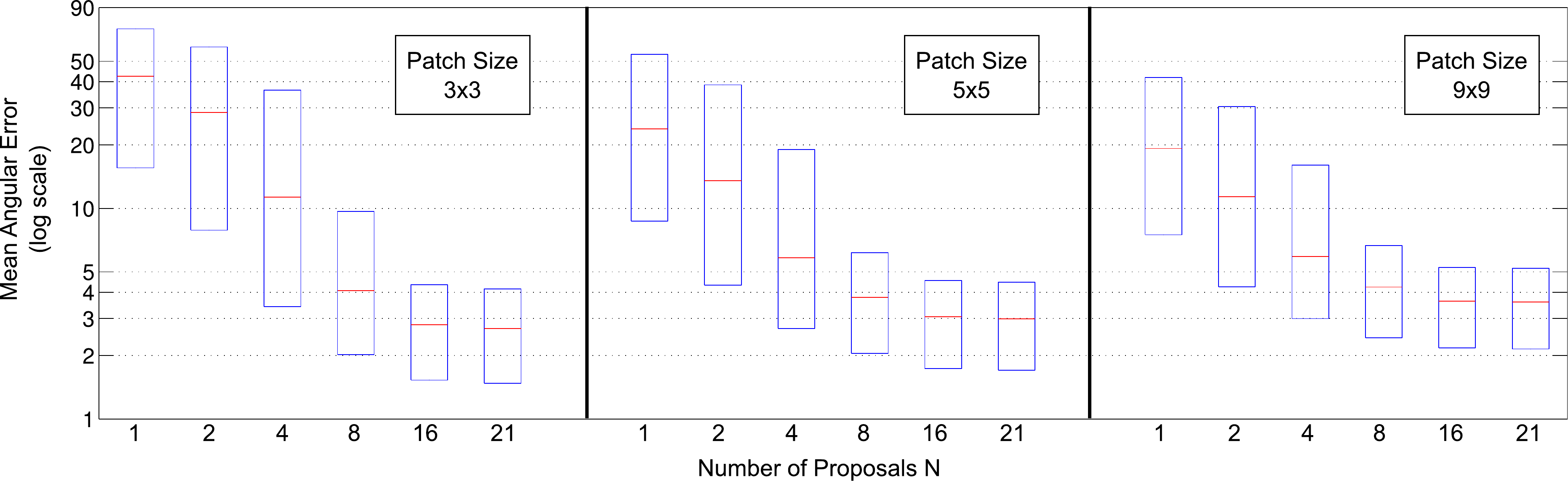}
  \caption{Local shape accuracy. We show quantiles (25\%, median, 75\%) of each patch's mean normal angular error, for the best estimate amongst the $N$ most-likely shape proposals for each patch, for different values of $N$ and for different patch sizes. The quantiles for $N=1$ correspond to making a hard decision at each patch, and errors for $N=21$ correspond to the best estimate amongst the full set of proposals.}
  \label{fig:boxplot}
\end{figure*}

We perform a quantitative evaluation of accuracy using all overlapping patches
from all six random surfaces and for three different patch-sizes (roughly
$80k-90k$ patches in total for each size). We are interested in knowing: (i) how
often the veridical shape is among the set of shape proposals $a^j$ at a patch;
and (ii) whether the cost $D^j$ is an informative metric for determining which
proposals $a^j$ are most accurate. To this end, for each image patch in the
evaluation set, we sort the proposed quadratic shapes according to their
likelihood costs $D^j$, compute for each proposal the mean angular difference
between its surface normals the veridical ones, and record for increasing values
$N=\{1,\ldots J=21\}$ the lowest mean angular error among the $N$ most-likely
shape proposals. Figure~\ref{fig:boxplot} shows the statistics of these errors
across all test patches for increasing values of $N$. Although the most-likely
shape proposal (\ie $N=1$) is often reasonably close to true shape, the error
quantiles decrease significantly more as we consider larger sets of likely
proposals. This emphasizes the value of maintaining full distributions of local
shape as a mid-level scene representation, as opposed to ``over-committing'' to
only one (often sub-optimal) shape proposal for each patch through a process of
hard local decision-making.

Figure~\ref{fig:boxplot} also provides some insight about the effect of patch-size, and it shows that patches at multiple scales tend to be complimentary. Smaller patches are more likely to have lower errors when considering the full set of proposals ($N=J=21$), since the veridical shape is much more likely to be exactly quadratic at smaller scales. But, as evidenced by the relatively smaller error quantiles for lower values of $N$, larger patches tend to be more informative, with their likelihoods $D^j$ being better predictors of \emph{which} of the proposals $a^j$ is the true one.

\section{Surface reconstruction}
\label{sec:reconstruction}

To demonstrate the utility of our theory and local distributions for higher-level scene analysis, we consider the application of reconstructing object-scale surface shape when the light $l$ is known. The local representations provide concise summaries of the shape information available in each image patch, and they do this without ``over-committing'' to any one local explanation. This allows us to achieve reliable performance with very a simple algorithm for global reasoning that infers object-scale shape
through simple iterations between: 1) choosing one likely shape proposal for each local patch; and 2) fitting a global smooth surface to the set of chosen per-patch proposals.

Formally, our goal is to estimate the depth map $Z(x,y)$ from an observed intensity image $I(x,y)$, with known lighting $l$ and under the assumption that the surface is predominantly texture-less and diffuse, \ie, the shading equation \eqref{eq:measurement} holds at most pixels. We first compute local distributions as described in the previous section, by dividing the surface into a mosaic of overlapping patches of different sizes. We let $p \in \{1,\ldots P\}$ index all patches (across different patch-sizes), with $\Omega_p$ corresponding to the pixel locations, and $\{a_p^j,D_p^j\}_j$ denoting the local shape proposals and distribution for each patch $p$. 

In addition to the $J$ proposals at each patch, we use an approach similar to \cite{coles2012hapecollage} and include a dummy proposal $\{a_p^{J+1}=\phi,D_p^{J+1}=D_\phi\}$ in the distribution for every patch. This serves to make the surface reconstruction robust to outliers, such as when the local patch deviates significantly from a quadratic approximation (\eg sharp edges or depth discontinuities), or when the observed intensities vary from the diffuse model in \eqref{eq:measurement}, \eg specularities, shadows, or albedo variations due to texture.

We formulate the reconstruction problem as simultaneously finding a depth estimate $Z$ and a labeling $L_p \in \{1,\ldots J+1\}, \forall p$ that minimize the cost function:
\begin{equation}
  \label{eq:reconstruct_cost}
  C(Z,\{L_p\},\lambda) = \sum_{p=1}^P \left(\lambda D_p^{L_p}+  \sum_{(x,y)\in\Omega_p} \delta(Z,a^{L_p}_p,x,y) \right),
\end{equation}
where $\lambda$ is a scalar weight, and $\delta$ measures the agreement between the local shape proposal $a^{L_p}_p$ and $Z$ at $(x,y)$:
\begin{equation}
  \label{eq:agreement}
  \delta(Z,a,x,y) = \left\|\begin{array}{c}\nabla_x Z(x,y) - n_x(x,y;a)\\\nabla_yZ(x,y) - n_y(x,y;a)\end{array}\right\|^2,
\end{equation}
if $a\neq \phi$, and $0$ otherwise. 

\begin{figure*}[!t]
\centering
\includegraphics[width=\textwidth]{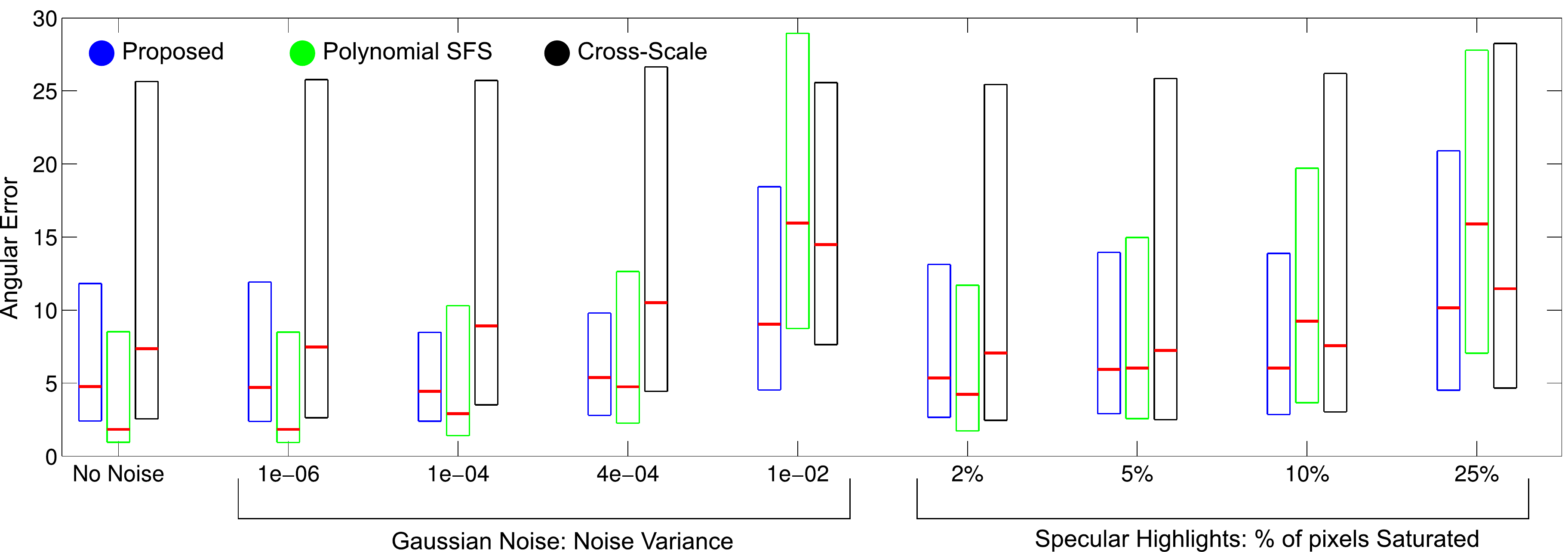}
\caption{Surface reconstruction accuracy for different methods on synthetic images of random surfaces. Shown here are quantiles (25\%, median, 75\%) of angular errors of individual normals across all surfaces for images rendered with different degrees of additive Gaussian noise, and specular highlights.\label{fig:SRboxplot}}
\end{figure*}

We use an iterative algorithm to minimize the cost $C$, alternating updates to $Z$ and $\{L_p\}$ based on the current estimate of the other. Given the current estimate $Z^*$ of the depth map at each iteration, we  update the label $L_p$ of every patch independently (and in parallel):
\begin{equation}
  \label{eq:lupdate}
  L_p \leftarrow \argmin_{L \in \{1,\ldots J+1\}}~~\lambda D_p^L + \sum_{(x,y)\in\Omega_p} \delta(Z^*,a_p^L,x,y).
\end{equation}

Similarly, given a labeling $\{L_p^*\}$ (with corresponding shape proposals $\{a_p^*\}$),  we  compute the depth map $Z$ that minimizes the cost in \eqref{eq:reconstruct_cost} as:
\begin{align}
  \label{eq:zupdate}
  &Z \leftarrow \argmin_Z \sum_p \sum_{(x,y)\in\Omega_p} \delta(Z,a_p^{*},x,y)\notag\\
  &= \argmin_Z \sum_{x,y} w^*(x,y) \left\|\begin{array}{c}
    \nabla_x Z(x,y) - n_x^*(x,y)\\
    \nabla_y Z(x,y) - n_y^*(x,y)
  \end{array}\right\|^2,
\end{align}
where $w^*(x,y)$ is the number patches that include $(x,y)$ and have not been labeled as outliers, and $n_x^*(x,y),n_y^*(x,y)$ their mean normal estimates:
\begin{align}
  &\Omega^{-1}(x,y) = \{p: (x,y) \in \Omega_p, a^*_p \neq \phi\},\notag\\
  &w^*(x,y) = |\Omega^{-1}(x,y)|,\notag\\
  &n_x^*(x,y) = \frac{1}{w^*(x,y)}\sum_{p \in \Omega^{-1}(x,y)} n_x(x,y;a^*_p),\notag\\
  &n_y^*(x,y) = \frac{1}{w^*(x,y)}\sum_{p \in \Omega^{-1}(x,y)} n_y(x,y;a^*_p).
\end{align}
The computation in \eqref{eq:zupdate} could be carried out exactly and efficiently using the Frankot-Chellappa algorithm~\cite{frankot1988method} if all $w^*(x,y)$ were equal. But this is not the case since $w^*(x,y)$ will be lower near the boundary and in regions where some patches have been detected as outliers. Nevertheless, we find that \cite{frankot1988method} provides an acceptable approximation in these cases. We use \cite{frankot1988method} throughout the alternating iterations until $Z$ and $\{L_p\}$ converge\footnote{We simply set $n_x^*(x,y)=n_y^*(x,y)=0$ when $w^*(x,y)=0$.}, and then we run a limited number of additional iterations using conjugate-gradient to compute step \eqref{eq:zupdate} exactly.

To speed up convergence, we smooth the estimate of $Z$ in the first few
iterations with a Gaussian filter with variance $\sigma$ and set
$\lambda'=\sigma^2\lambda$, starting from an initial value $\sigma_0$ that is
decreased by a constant factor $\sigma_f$ till it reaches $1$ (at which point we
stop smoothing). We also initially run the algorithm over only the valid
proposals at each patch till convergence, and then introduce the dummy proposal
$\phi$. We set the parameters $\lambda$ and $D_\phi$ automatically based on the
input distributions---$\lambda$ is set to $1/4$th the reciprocal of the mean of
the differences between the minimum and median likelihood costs across all
patches at the smallest scale, and $D_\phi$ is set to $10\lambda^{-1}$. The
reconstruction from local patch proposals takes 40 seconds on average on an
eight-core machine for $128\times 128$ images with local distributions at four
scales (not including the computation time for estimating local proposals, which
is reported in Sec.~\ref{sec:shape-proposal}).

\subsection{Evaluation}

We first quantitatively evaluate the proposed reconstruction algorithm, under known lighting, with the random surfaces described in Sec.~\ref{sec:local-evaluation}. We render images with different amounts of additive white Gaussian noise, as well as with specular highlights. For the latter, we use the Beckmann~\cite{beckmann} model and consider different values of ``surface smoothness'' to get images with increasing numbers of saturated pixels. We mask out pixels that are saturated during estimation, but note that many nearby unclipped pixels will also include a non-zero specular component that violates the diffuse shading model.

Figure~\ref{fig:SRboxplot} summarizes the performance---using local distributions of $3\times 3$, $5\times 5$, $9\times 9$, and $17\times 17$ overlapping patches---and compares it to two state-of-the-art methods. The first is the iterative algorithm proposed by Ecker and Jepson~\cite{EckerJ:2010} (labeled ``Polynomial SFS''). The second (labeled ``Cross-scale'') is the shape from shading component of the SIRFS method~\cite{BarronM:2013}, \ie, where we treat the light and shading-image as given, and do not use contour information. The cross-scale method uses an over-complete, multi-scale representation of the global depth map and minimizes the rendering error along with the likelihood of the recovered shape under a prior model. For both methods we use implementations provided by their authors, and for the cross-scale method, we use the author-provided prior parameters that were trained on the MIT intrinsic image database~\cite{GrosseJAF:2009}.\footnote{We also evaluated the cross-scale method with a prior trained on the random surfaces, but this did not improve performance.}

\begin{figure}[!t]
\centering
\includegraphics[width=\columnwidth]{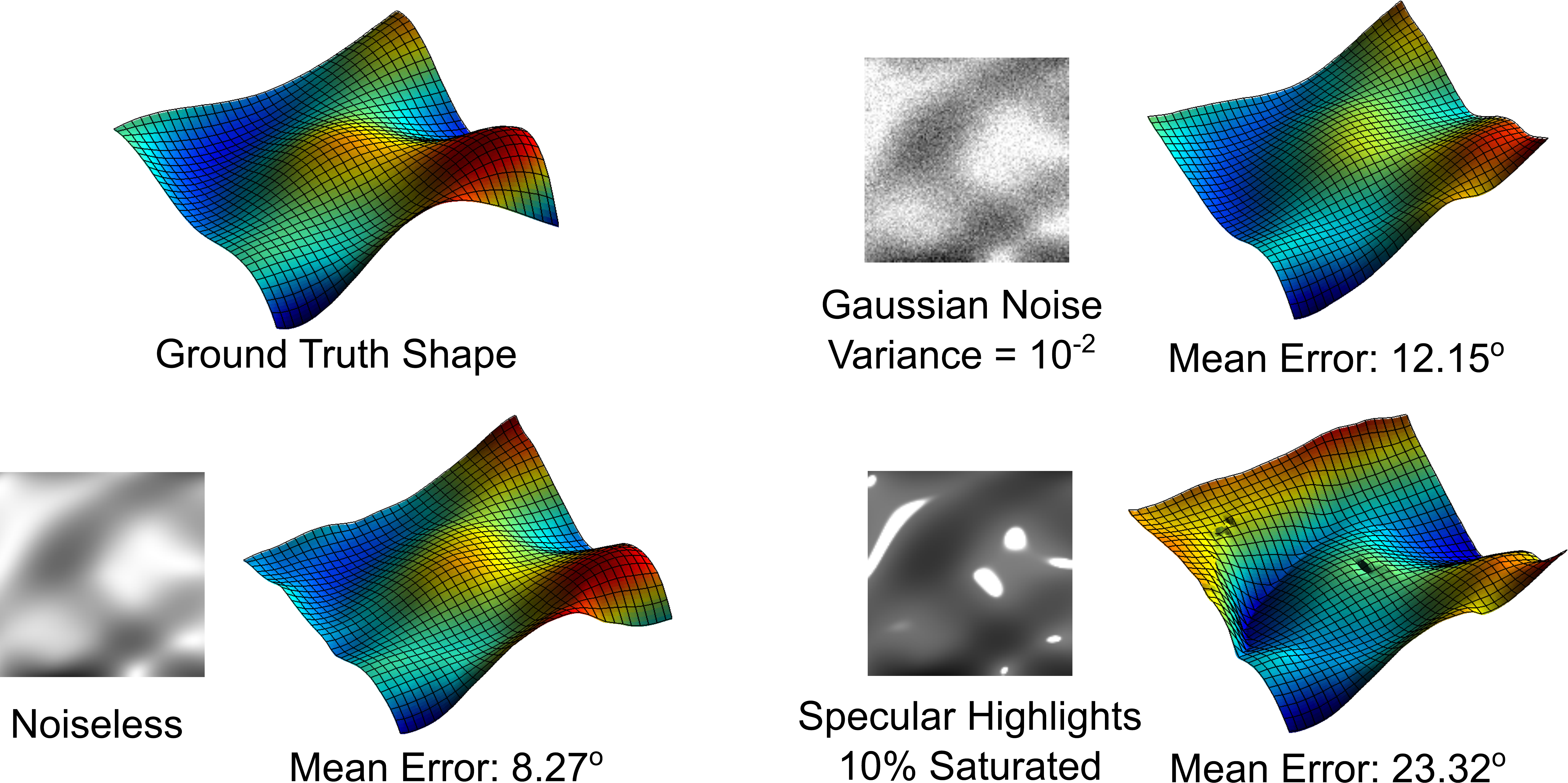}
\caption{Reconstructions by the proposed method on different rendered images of a synthetic surface.\label{fig:SRex}}
\end{figure}

We see from Fig.~\ref{fig:SRboxplot} that while the polynomial SFS method performs the best in the noiseless case, the proposed algorithm is more robust to both Gaussian noise and the structured artifacts from specular highlights. The cross-scale method is also reasonably robust to these effects due to its use of a shape prior, but in general has higher errors. Figure~\ref{fig:SRex} provides example reconstructions for the proposed method for one surface---the full set of reconstructions are available at \cite{page}.

Next, we evaluate all algorithms using photographs of seven relatively-diffuse
objects, captured with a Canon EOS 40D camera under directional lighting, with
two chrome spheres in the scene to measure light direction. These photographs
contain non-idealities such as mutual illumination, self-shadowing, and slight
texture. For each object under a fixed viewpoint, we took twenty images with
varying light directions, with which we can recover the normal vectors as well
as depth map by photometric stereo to a high accuracy. We use this recovered
shape as ground truth for our evaluation. All the captured images, calibration
information, and recovered normal and depth maps are available on our project
page~\cite{page}.

\begin{figure*}
\centering
\begin{tabular}{c|c|c|c}

Input Image & Proposed & Polynomial SFS & Cross-Scale \\

\vspace{-0.5em}
\raisebox{1em}{\includegraphics[width=0.12\textwidth]{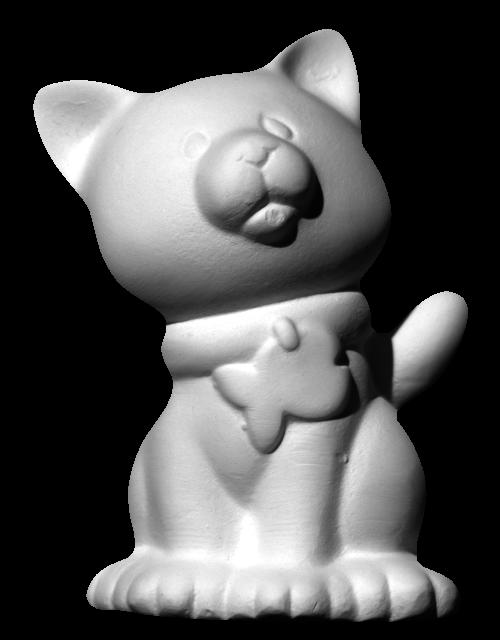}} &
\raisebox{0.0em}{\includegraphics[width=0.11\textwidth]{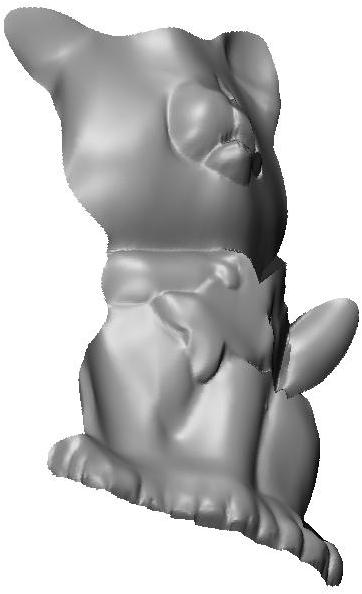}}
\raisebox{0.5em}{\includegraphics[width=0.13\textwidth]{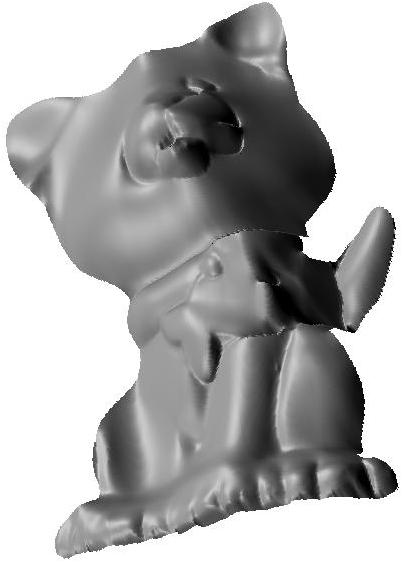}} &
\raisebox{0.0em}{\includegraphics[width=0.10\textwidth]{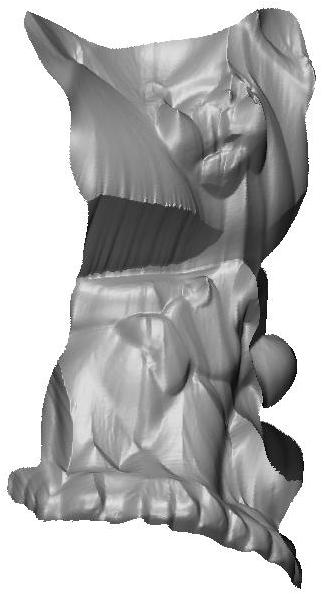}}
\raisebox{0.5em}{\includegraphics[width=0.14\textwidth]{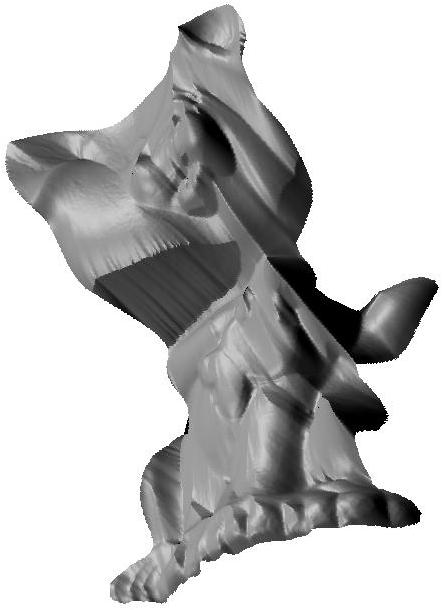}} &
\raisebox{0.0em}{\includegraphics[width=0.115\textwidth]{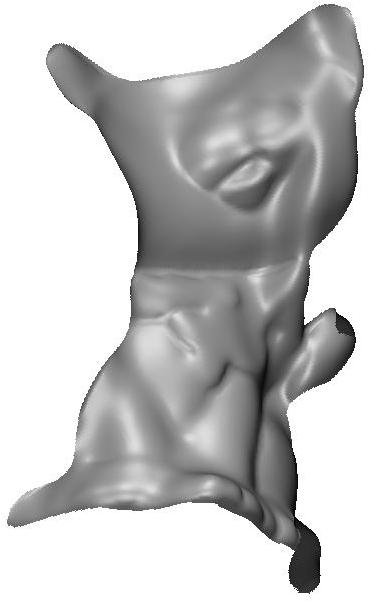}}
\raisebox{0.5em}{\includegraphics[width=0.125\textwidth]{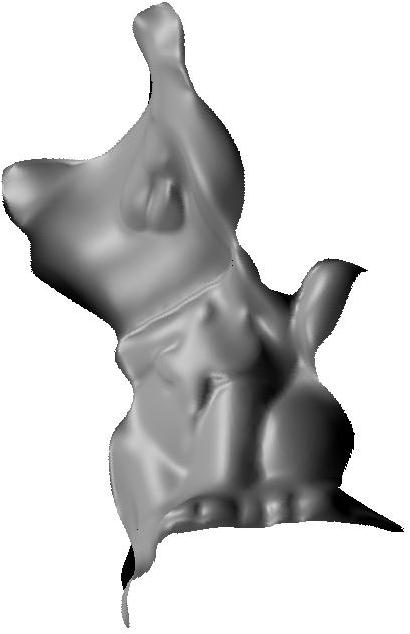}} \\
\scriptsize{Resolution $640\times 500$} &
\scriptsize{Median Angular Error $14.83^\circ$} &
\scriptsize{Median Angular Error $24.81^\circ$} &
\scriptsize{Median Angular Error $20.02^\circ$} \\

\vspace{-0.5em}
\raisebox{1em}{\includegraphics[width=0.12\textwidth]{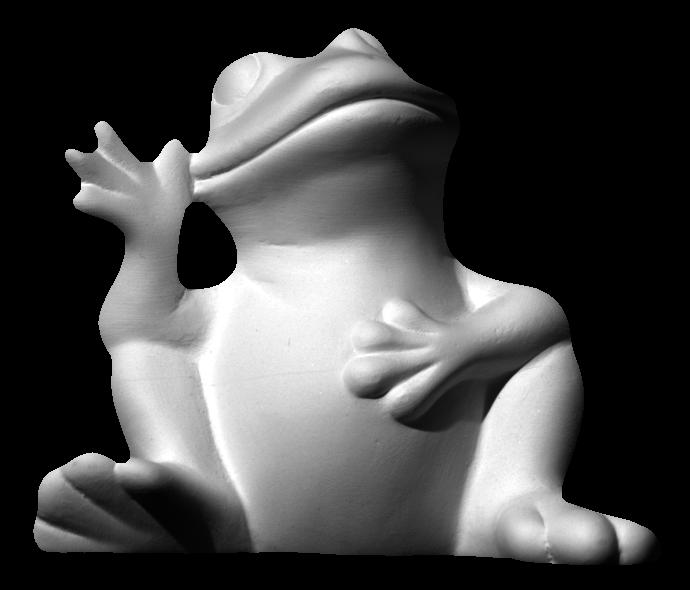}} &
\raisebox{0.0em}{\includegraphics[width=0.12\textwidth]{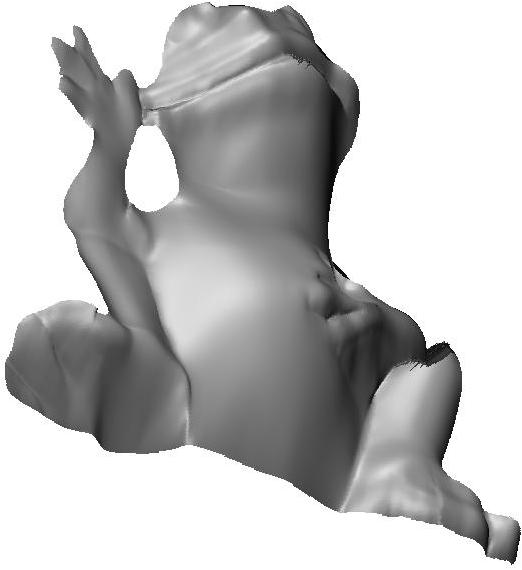}}
\raisebox{0.5em}{\includegraphics[width=0.12\textwidth]{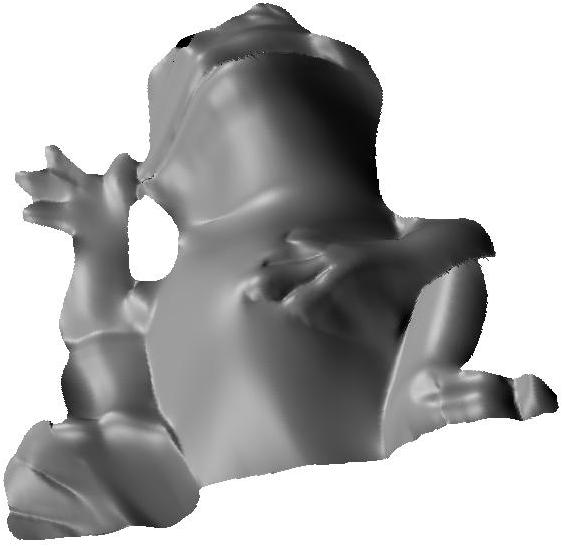}} &
\raisebox{0.0em}{\includegraphics[width=0.11\textwidth]{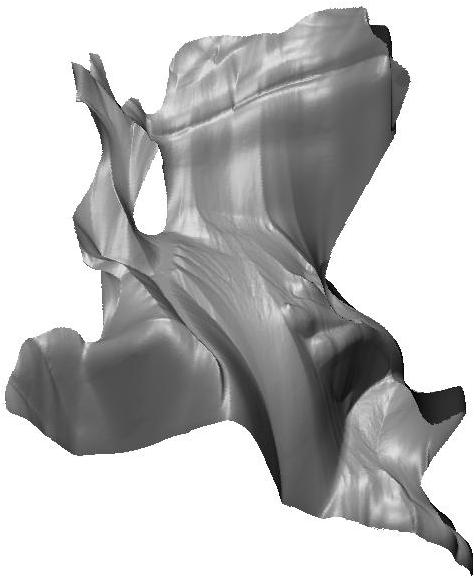}}
\raisebox{0.5em}{\includegraphics[width=0.13\textwidth]{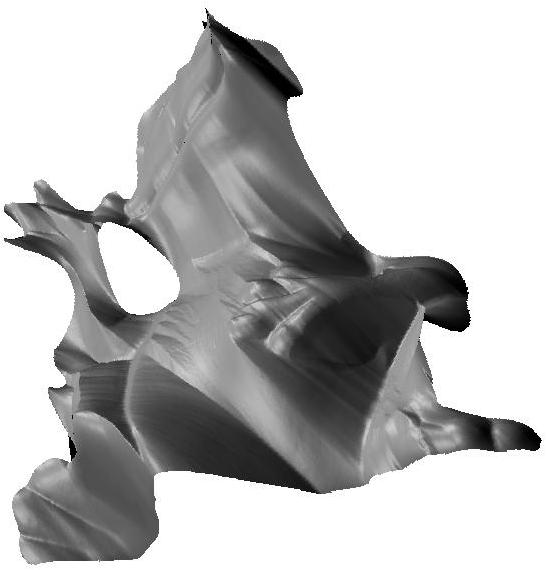}} &
\raisebox{0.0em}{\includegraphics[width=0.11\textwidth]{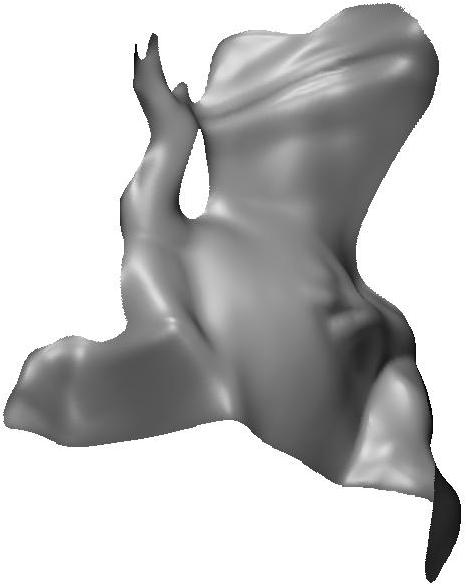}}
\raisebox{0.5em}{\includegraphics[width=0.13\textwidth]{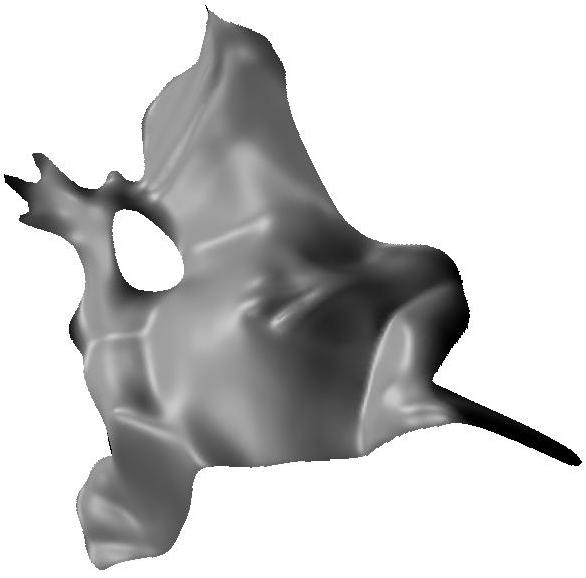}} \\
\scriptsize{Resolution $590\times 690$} &
\scriptsize{Median Angular Error $11.80^\circ$} &
\scriptsize{Median Angular Error $20.77^\circ$} &
\scriptsize{Median Angular Error $19.86^\circ$} \\

\vspace{-0.5em}
\raisebox{1em}{\includegraphics[width=0.12\textwidth]{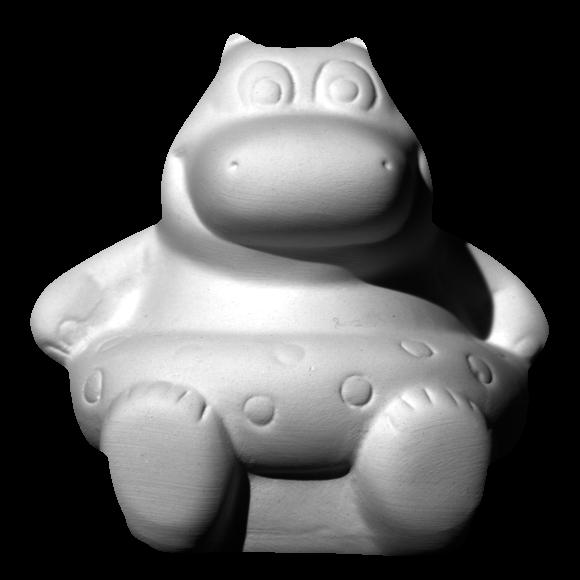}} &
\raisebox{0.0em}{\includegraphics[width=0.115\textwidth]{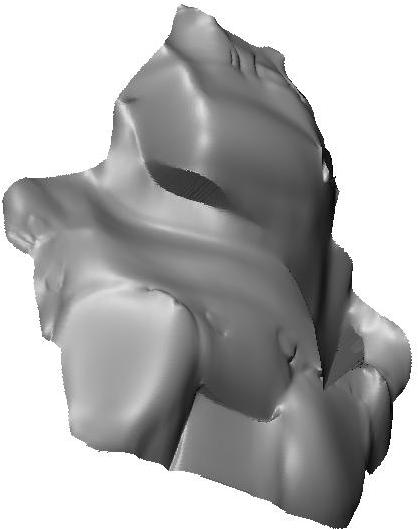}}
\raisebox{0.5em}{\includegraphics[width=0.125\textwidth]{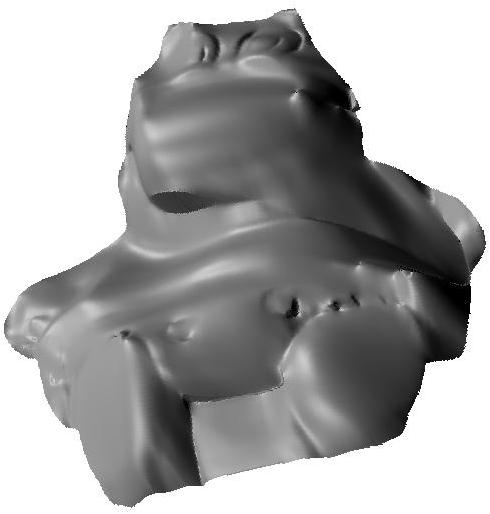}} &
\raisebox{0.0em}{\includegraphics[width=0.11\textwidth]{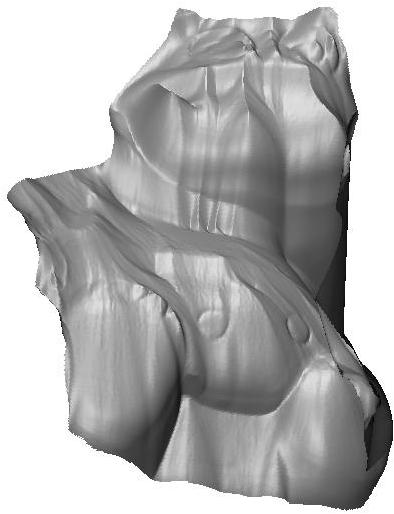}}
\raisebox{0.5em}{\includegraphics[width=0.13\textwidth]{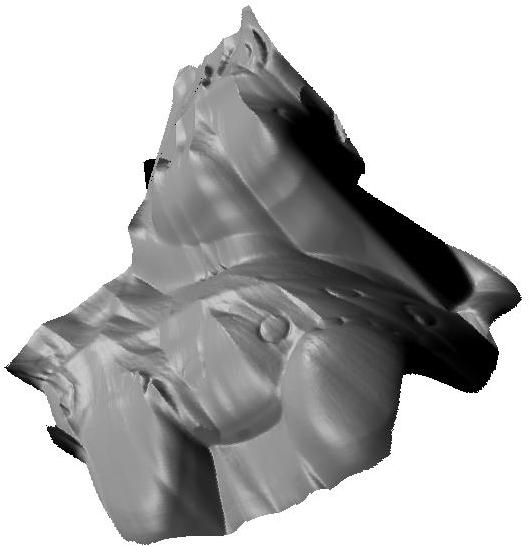}} &
\raisebox{0.0em}{\includegraphics[width=0.11\textwidth]{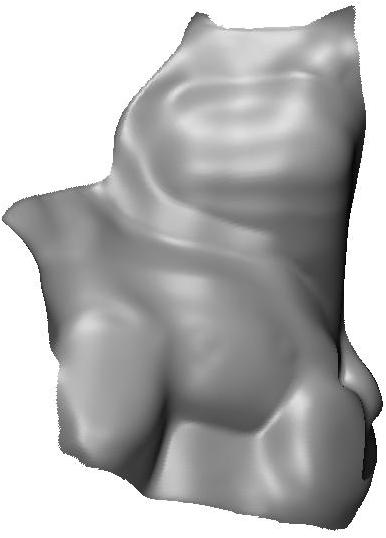}}
\raisebox{0.5em}{\includegraphics[width=0.13\textwidth]{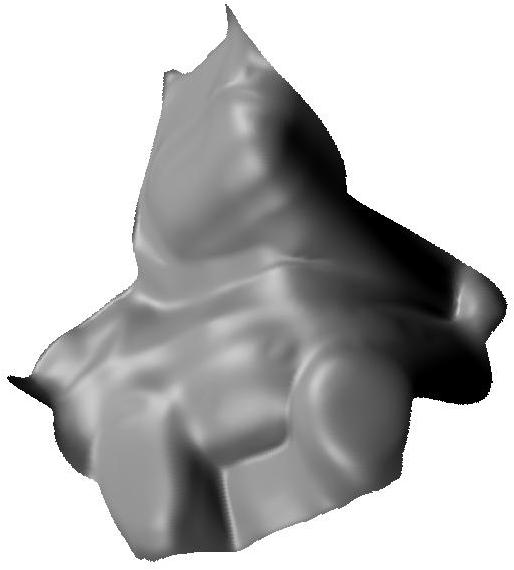}} \\
\scriptsize{Resolution $580\times 580$} &
\scriptsize{Median Angular Error $20.25^\circ$} &
\scriptsize{Median Angular Error $17.50^\circ$} &
\scriptsize{Median Angular Error $21.00^\circ$} \\

\vspace{-0.5em}
\raisebox{1em}{\includegraphics[width=0.12\textwidth]{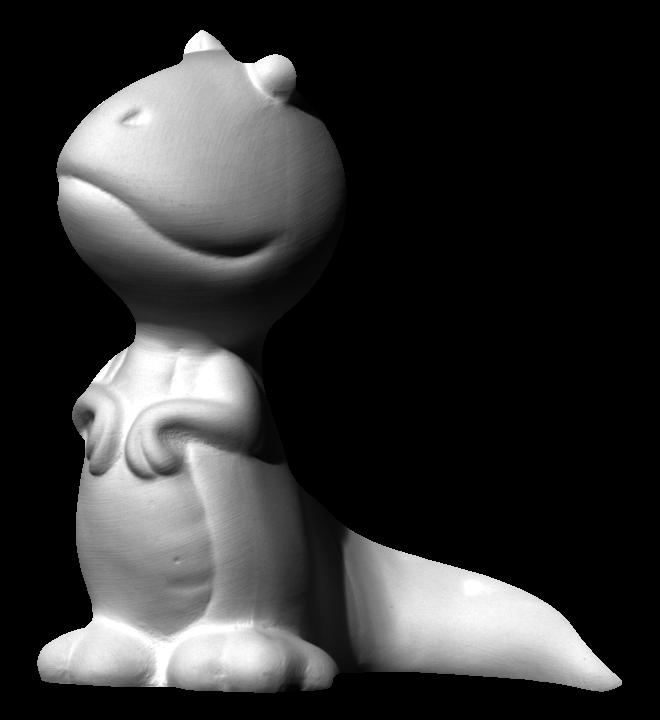}} &
\raisebox{0.0em}{\includegraphics[width=0.115\textwidth]{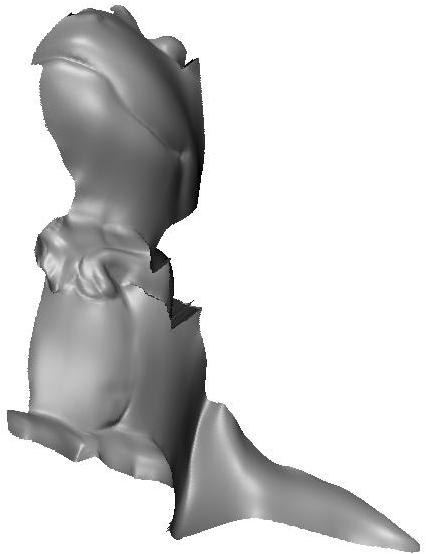}}
\raisebox{0.5em}{\includegraphics[width=0.125\textwidth]{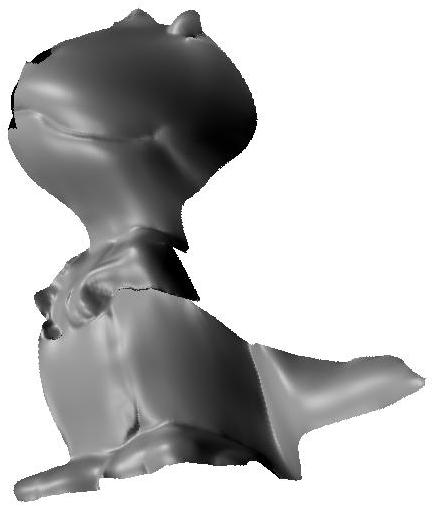}} &
\raisebox{0.0em}{\includegraphics[width=0.085\textwidth]{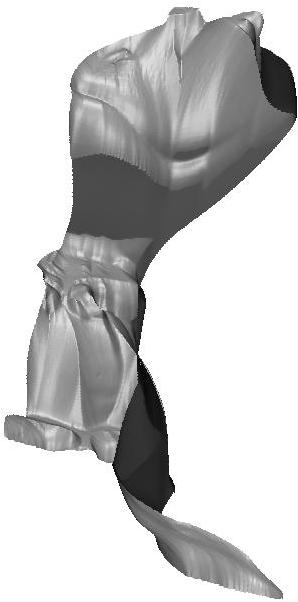}}
\raisebox{0.5em}{\includegraphics[width=0.145\textwidth]{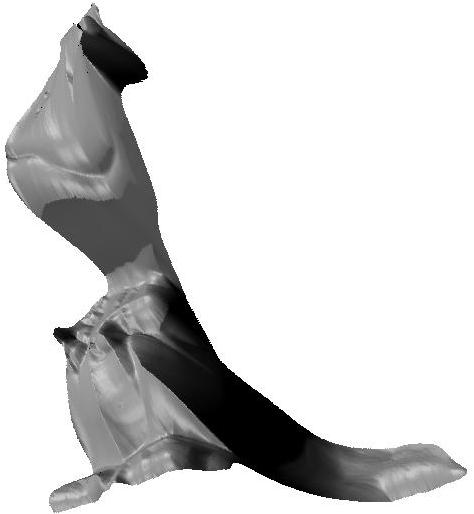}} &
\raisebox{0.0em}{\includegraphics[width=0.095\textwidth]{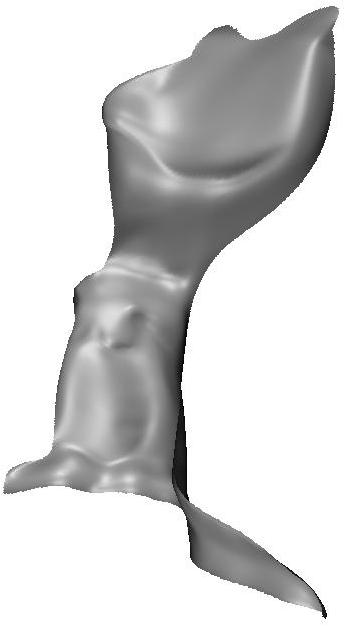}}
\raisebox{0.5em}{\includegraphics[width=0.135\textwidth]{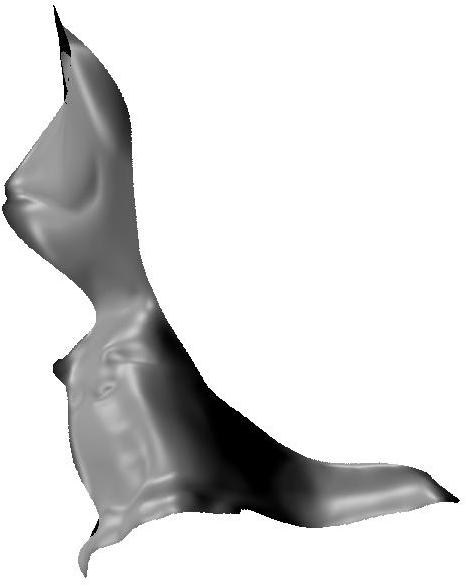}} \\
\scriptsize{Resolution $720\times 660$} &
\scriptsize{Median Angular Error $12.70^\circ$} &
\scriptsize{Median Angular Error $22.33^\circ$} &
\scriptsize{Median Angular Error $23.26^\circ$} \\

\vspace{-0.5em}
\raisebox{1em}{\includegraphics[width=0.12\textwidth]{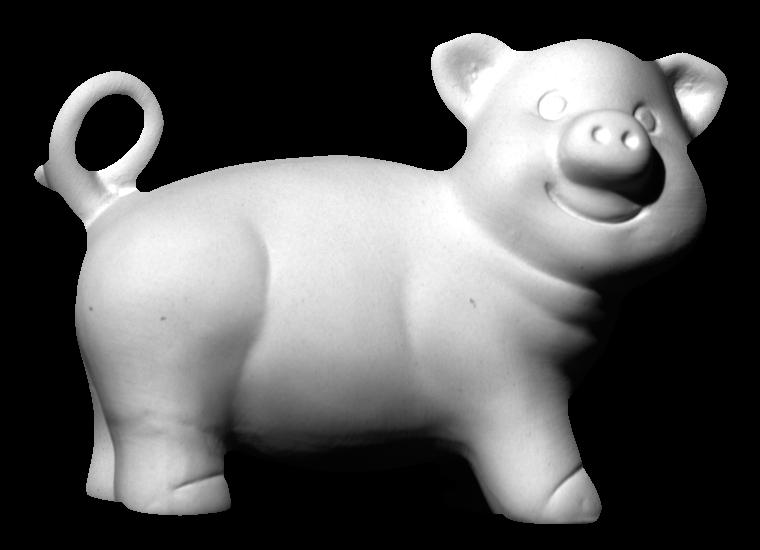}} &
\raisebox{0.0em}{\includegraphics[width=0.12\textwidth]{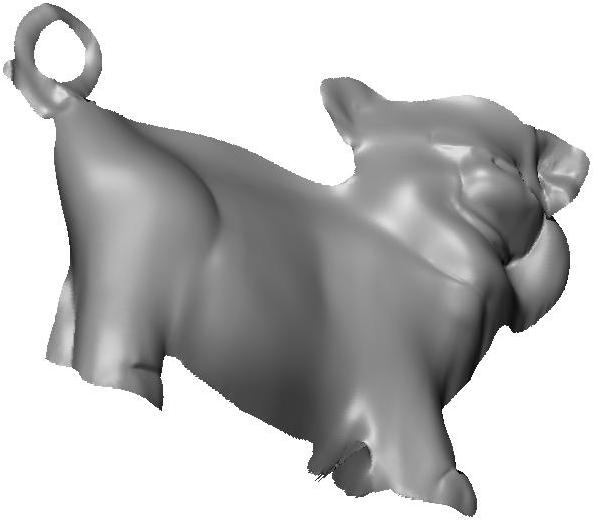}}
\raisebox{0.0em}{\includegraphics[width=0.12\textwidth]{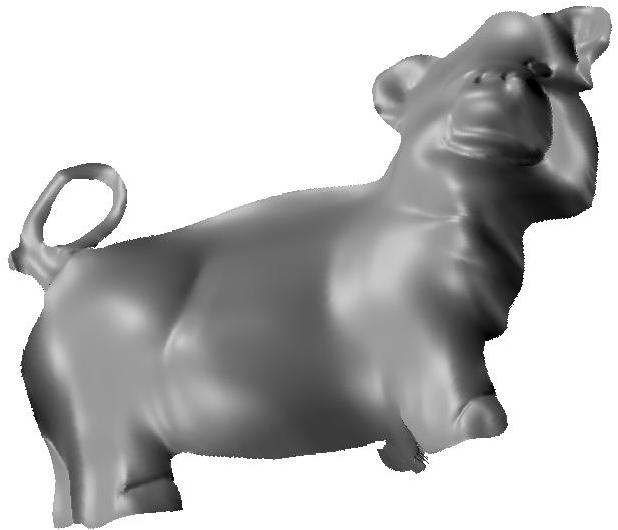}} &
\raisebox{0.0em}{\includegraphics[width=0.12\textwidth]{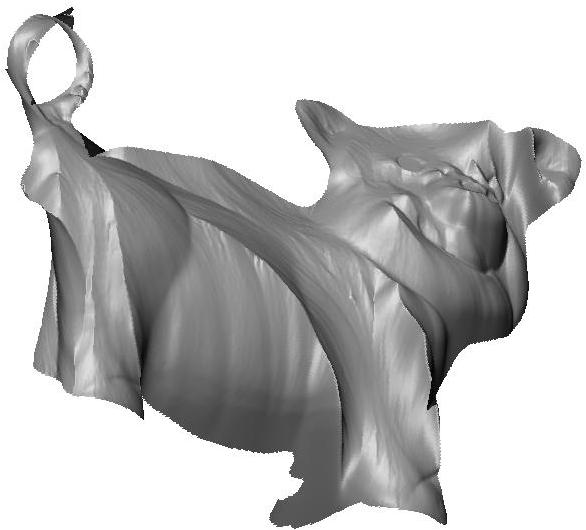}}
\raisebox{0.0em}{\includegraphics[width=0.12\textwidth]{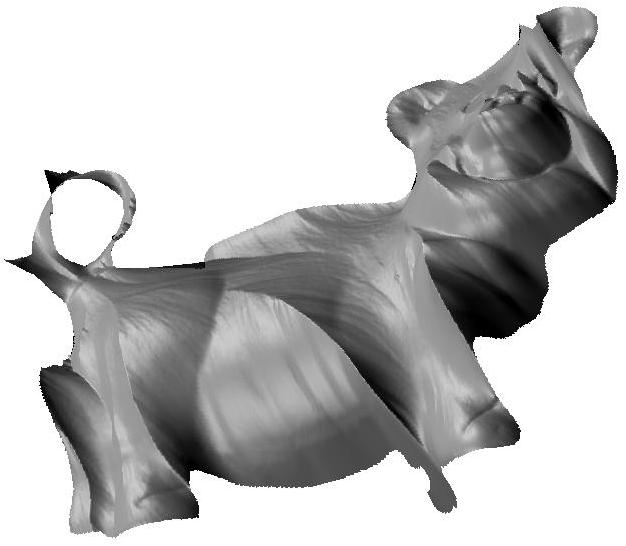}} &
\raisebox{0.2em}{\includegraphics[width=0.12\textwidth]{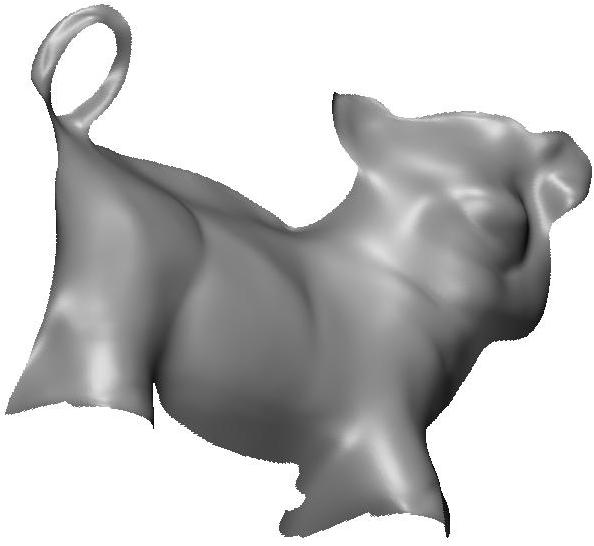}}
\raisebox{0.0em}{\includegraphics[width=0.12\textwidth]{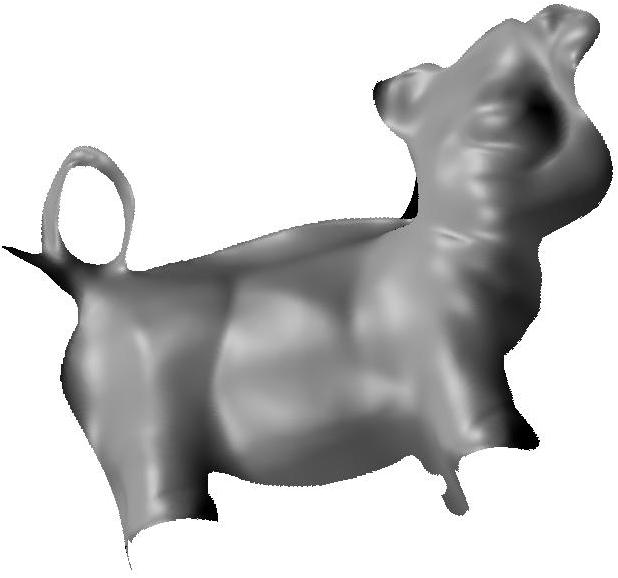}} \\
\scriptsize{Resolution $550\times 760$} &
\scriptsize{Median Angular Error $15.29^\circ$} &
\scriptsize{Median Angular Error $15.58^\circ$} &
\scriptsize{Median Angular Error $13.17^\circ$} \\

\vspace{-0.5em}
\raisebox{1em}{\includegraphics[width=0.12\textwidth]{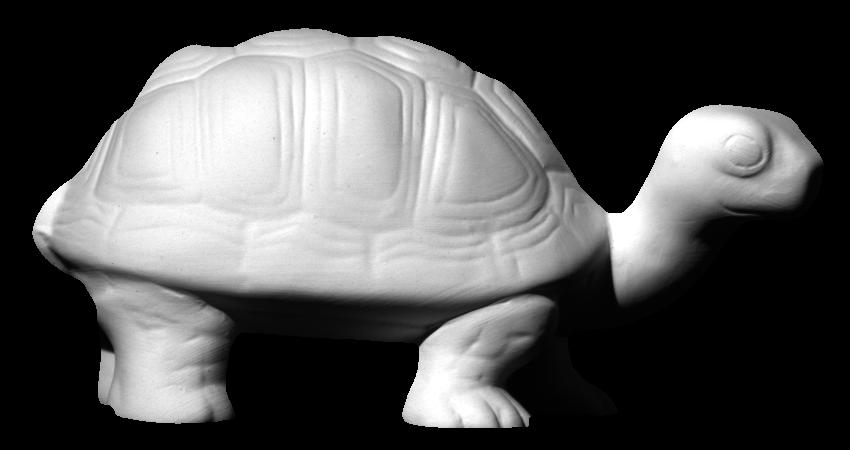}} &
\raisebox{0.0em}{\includegraphics[width=0.12\textwidth]{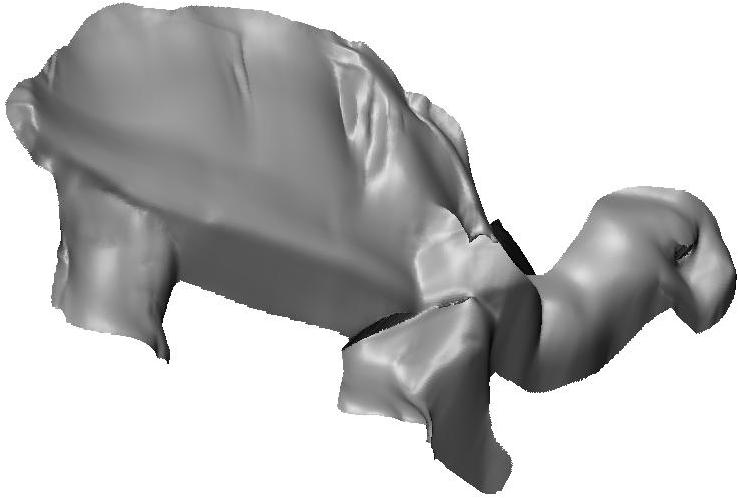}}
\raisebox{0.0em}{\includegraphics[width=0.12\textwidth]{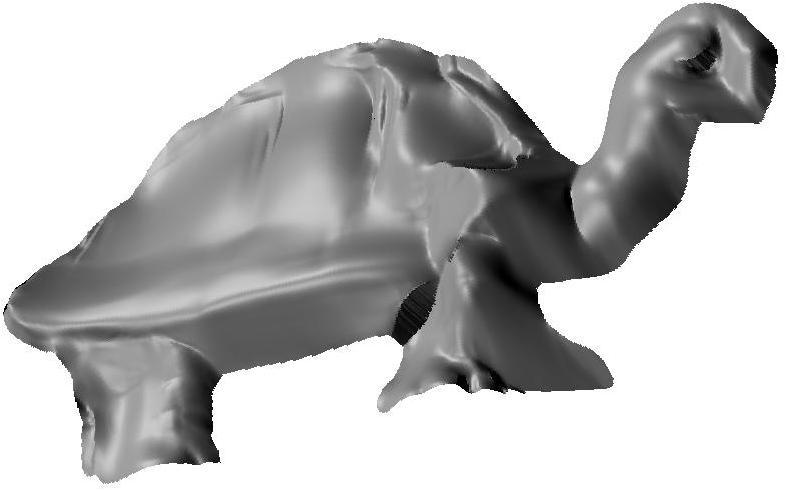}} &
\raisebox{0.0em}{\includegraphics[width=0.12\textwidth]{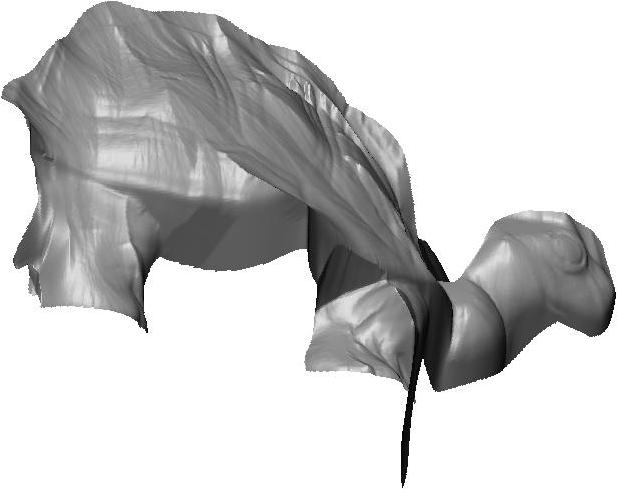}}
\raisebox{0.5em}{\includegraphics[width=0.12\textwidth]{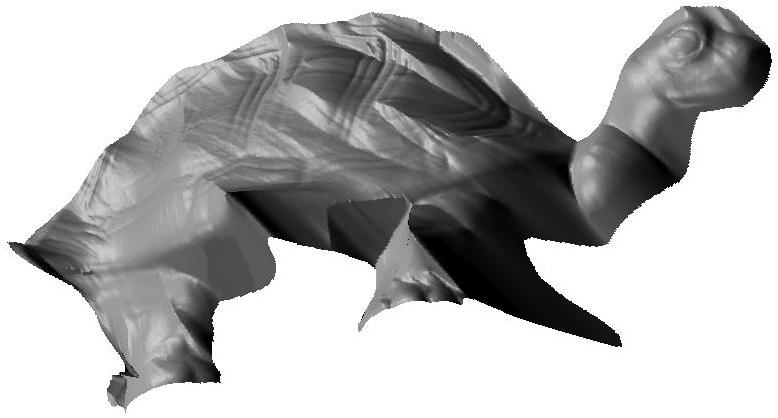}} &
\raisebox{0.0em}{\includegraphics[width=0.12\textwidth]{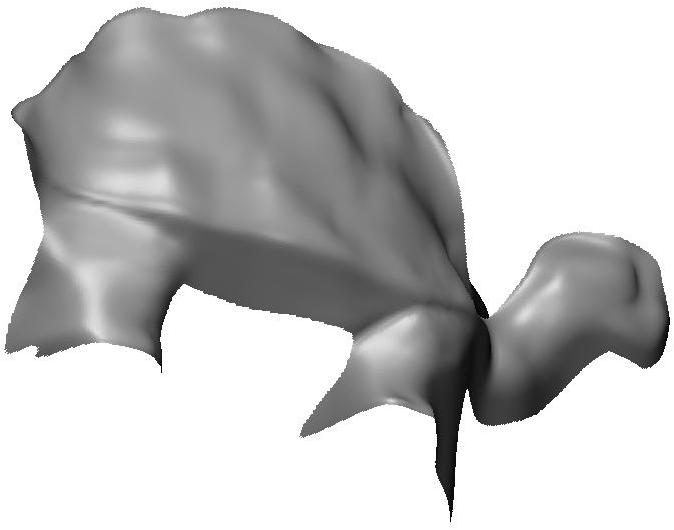}}
\raisebox{0.2em}{\includegraphics[width=0.12\textwidth]{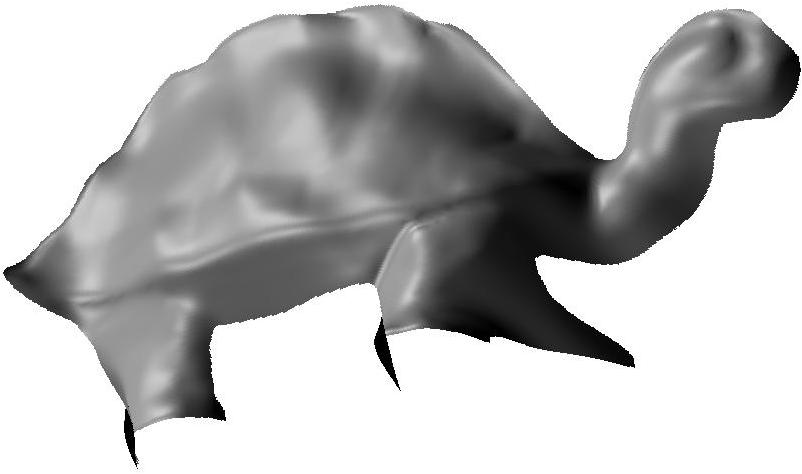}} \\
\scriptsize{Resolution $450\times 850$} &
\scriptsize{Median Angular Error $17.90^\circ$} &
\scriptsize{Median Angular Error $14.50^\circ$} &
\scriptsize{Median Angular Error $11.96^\circ$} \\

\vspace{-0.5em}
\raisebox{1em}{\includegraphics[width=0.12\textwidth]{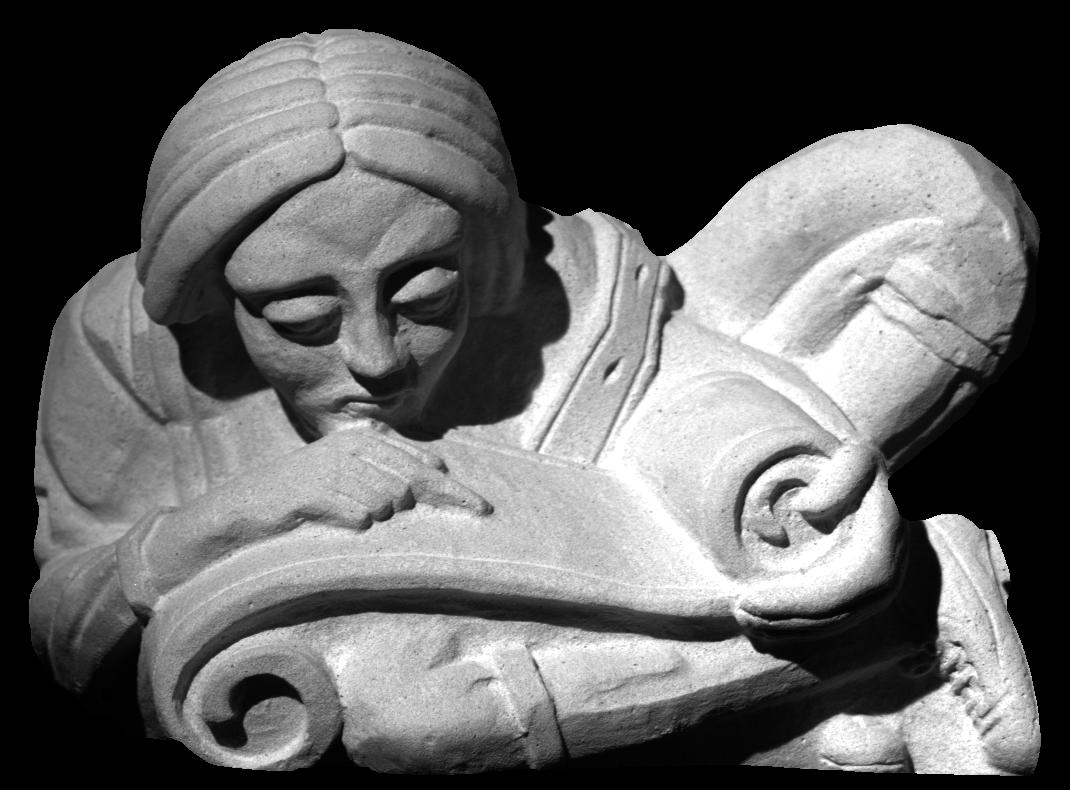}} &
\raisebox{0.0em}{\includegraphics[width=0.12\textwidth]{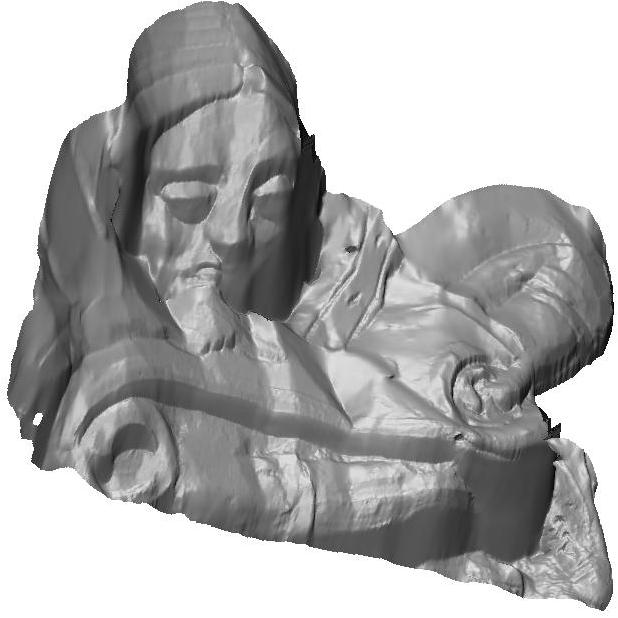}}
\raisebox{0.5em}{\includegraphics[width=0.12\textwidth]{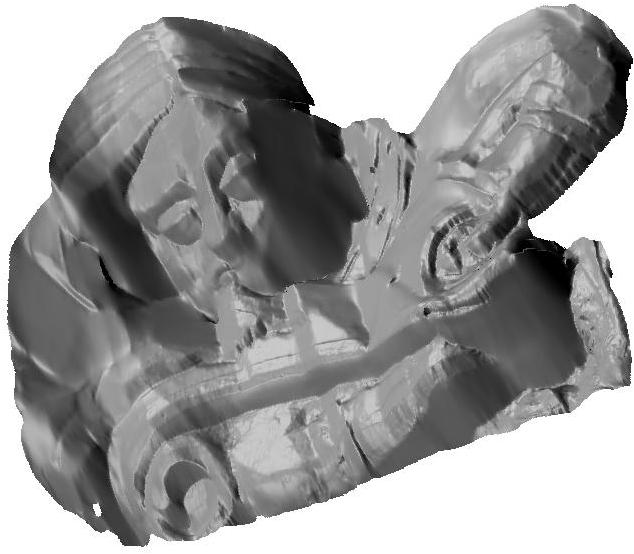}} &
\raisebox{0.0em}{\includegraphics[width=0.11\textwidth]{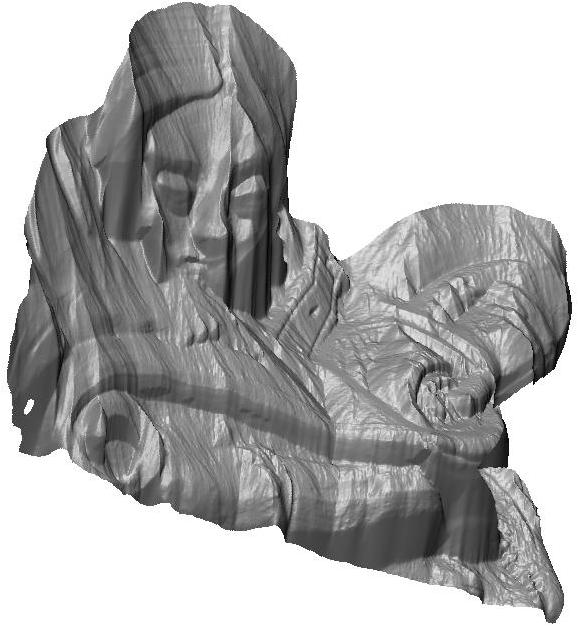}}
\raisebox{0.5em}{\includegraphics[width=0.13\textwidth]{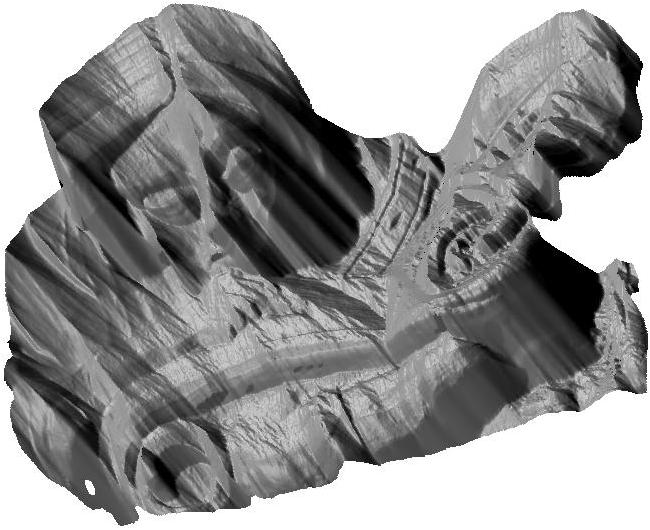}} &
\raisebox{0.0em}{\includegraphics[width=0.12\textwidth]{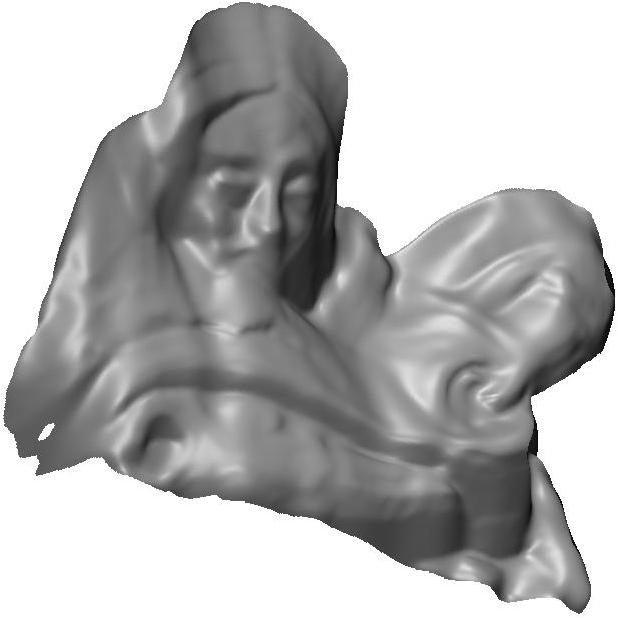}}
\raisebox{0.5em}{\includegraphics[width=0.12\textwidth]{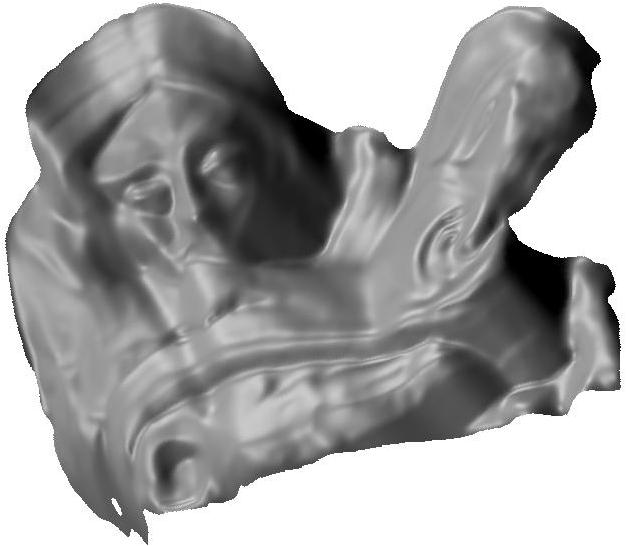}} \\
\scriptsize{Resolution $790\times 1070$} &
\scriptsize{Median Angular Error $28.13^\circ$} &
\scriptsize{Median Angular Error $29.21^\circ$} &
\scriptsize{Median Angular Error $25.80^\circ$} \\

\end{tabular}\vspace{0.5em}

\caption{Surface reconstruction on real captured data. We show two novel view points for each reconstruction, and the median angular error between estimated surface normal vectors and ground truth surface normal vectors. A more interactive visualization is available at \cite{page}.\label{fig:reconstruction-real}}
\end{figure*}

For each object, we choose a single image as input to evaluate the performance
of different SFS frameworks. The 99th percentile intensity value of the image is
assumed to correspond to the albedo times light intensity and used for image
normalization; and since these images are larger, we use local distributions at
two additional patch-scales: $33\times 33$ and $65\times 65$. The surfaces
reconstructed using the different methods and measured light direction are shown
in Fig.~\ref{fig:reconstruction-real} along with median angular error values. We find that in most cases, the proposed algorithm provides a better reconstructions of object-scale shape than the baselines.

\section{Discussion}

Our theoretical analysis shows that in an idealized quadratic world, local shape can be recovered uniquely in almost every local image patch, without the use of singular points, occluding contours, or any other external shape information. Beyond this idealized world, our evaluations on synthetic and captured images suggest that one can infer, efficiently and in parallel, concise multi-scale local shape distributions that are accurate and useful for global reasoning.

There are many viable directions for interesting future work. Foremost among these is the joint estimation of shape, lighting, and albedo. The reconstruction algorithms proposed in this paper are limited to the case when lighting is known, but the uniqueness results in Sec.~\ref{sec:generallight} suggest that simultaneous reconstruction of shape and lighting may also be possible. Theorem~\ref{prop:4-sol} tells us that, in an idealized quadratic world, there are generically four lights $l$ that can explain each local patch, and that these quadruples of possible lights will vary from patch to patch according to the directions of each patch's Hessian eigenvectors. Intuitively, one might infer the true light (along with its reflection across the view, which is always equally-likely) as the one that is common to all or most of the per-patch quadruples.\footnote{We have experimented with a direct implementation of this intuition that does a brute-force search only on lighting direction, assuming a known constant light-strength and albedo, and with pooling local estimates without considering consistency or noise. This method worked reasonably well in many cases, but was computationally expensive and not entirely robust.} Practically speaking, it is likely that for a reconstruction algorithm to handle unknown lighting, it will need to jointly reason about shape, lighting, and varying albedo, in the same spirit as Barron and Malik~\cite{BarronM:2013}; and that such reasoning will benefit from an analysis of the joint ambiguities that are induced by noise and non-quadratic shape, similar to what was done for shape alone in Sections~\ref{sec:noise_instability} and \ref{sec:stochastic}.

Also, while we provide a means to extract a single estimate of the global surface from local shape distributions, one could also imagine using reasoning about consistency and outliers to allow the full distributions of neighboring patches to collaboratively refine themselves. This could be useful, for instance, when the object boundaries in a scene are not known a-priori. These refined local distributions may then be able to identify depth discontinuities in the scene, and help segment out individual objects for shape recovery.

Finally, it will be interesting to pursue combining our shading-based local distributions with complementary reasoning about contours, shading keypoints~\cite{forsyth98}, texture, gloss, shadows, and so on---treating these as additional cues for shape, as well as to better identify outliers to our smooth diffuse shading model. We also believe it is worth integrating these local shape distributions into processes for higher-level vision tasks such as pose estimation, object recognition, and multi-view reconstruction, where one can imagine additionally using top-down processing to aid local inference, for example by exploiting priors on local quadratic shapes that are based on object identity or scene category.


\clearpage

\noindent {\bf \large Supplementary Material}\\

{\small

In this supplementary section, we provide a proof for Lemma \ref{lem:AllA} from Sec.~\ref{sec:generallight}. As a reminder, the lemma is defined in terms of matrices $A \in \mathbb{R}^{3\times 3}$ which are related to the coefficient vectors as:
\begin{equation*}
    A=\left[\begin{array}{ccc}
      -2a_{1} & -a_{3} & -a_{4}\\
      -a_{3} & -2a_{2} & -a_{5}\\
      0 & 0 & 1
    \end{array}\right].\tag{\mbox{\ref{eq:Axdef}}}
\end{equation*}
The statement of the lemma itself is reproduced below.

~

\noindent {\bf Lemma \ref{lem:AllA}} : {\em Let $A$ and $\widetilde{A}$ correspond to two coefficient matrices of the form in \eqref{eq:Axdef}, and $l$ and $\widetilde{l}$ to two lighting vectors. If,
\begin{equation}
  \frac{\xvec^TA^Tll^TA\xvec}{{\xvec^TA^TA\xvec}} = \frac{\xvec^T\widetilde{A}^T\widetilde{l}\widetilde{l}^T\widetilde{A}\xvec}{{\xvec^T\widetilde{A}^T\widetilde{A}\xvec}}, \forall \xvec \in \Omega,\tag{\mbox{\ref{eq:alla2}}}
\end{equation}
Rank$(V_\Omega) = 15$, Rank$(A) \geq 2$, and $l^TA\xvec> 0, \forall \xvec \in \Omega$ (i.e., no point is in shadow), then
\begin{equation}
A^Tll^TA = \widetilde{A}^T\widetilde{l}\widetilde{l}^T\widetilde{A},~~A^TA = \widetilde{A}^T\widetilde{A}.\tag{\mbox{\ref{eq:alla}}}
\end{equation}
Moreover, if Rank$(A)=2$, then Rank$(\widetilde{A})=2$ and both $A$ and $\widetilde{A}$ have a common null space.
}

~

The expression in \eqref{eq:alla2} equates two rational forms in $x$. To prove the lemma, we will show that the equality holds for all $x$ if we have a sufficient number of non-degenerate locations in the patch. Then, we will show that the corresponding coefficients in the quadratic expressions in the numerator and denominator must be equal when they are of the form in \eqref{eq:Axdef} and the conditions of Lemma~\ref{lem:AllA} are met, essentially ruling out the possibility of a common factor or scaling term. To this end, we introduce another lemma, with proof, and then present the proof of Lemma~\ref{lem:AllA}.

\begin{lemma}
\label{lem:PQ}
Let $P,Q,\widetilde{P},\widetilde{Q}\in\mathbb{R}^{3\times3}$ be
symmetric matrices. Then, $P=t\widetilde{P}$ and $Q=t\widetilde{Q}$, where $t \neq 0$ is a constant scalar, if
\begin{equation}
\frac{\xvec^{T}P\xvec}{\xvec^{T}Q\xvec}=\frac{\xvec^{T}\widetilde{P}\xvec}{\xvec^{T}\widetilde{Q}\xvec},\quad\forall\xvec\in\Omega,
\label{eq:xPx/xQx}
\end{equation}
Rank($V_\Omega$) = 15, and,

~

\noindent {\bf Case 1:} All of the following conditions are satisfied:
\begin{align}
&q_{11}\neq0,\quad q_{22}\neq0,\quad q_{33}\neq0,\label{eq:q-neq0}\\
&4(p_{11}q_{12}-p_{12}q_{11})(p_{22}q_{12}-p_{12}q_{22})~+~\qquad\nonumber\\
&\qquad\qquad\qquad\qquad\qquad\qquad(p_{11}q_{22}-p_{22}q_{11})^{2}\neq0,\label{eq:pq12}\\
&4(p_{11}q_{13}-p_{13}q_{11})(p_{33}q_{13}-p_{13}q_{33})~+~\nonumber \\
&\qquad\qquad\qquad\qquad\qquad\qquad(p_{11}q_{33}-p_{33}q_{11})^{2}\neq0,\label{eq:pq13}\\
&4(p_{22}q_{23}-p_{23}q_{22})(p_{33}q_{23}-p_{23}q_{33})~+~\qquad\nonumber \\
&\qquad\qquad\qquad\qquad\qquad\qquad(p_{22}q_{33}-p_{33}q_{22})^{2}\neq0.\label{eq:pq23}
\end{align}

\noindent{\bf Case 2:} All of the following conditions are satisfied:
\begin{align}
&p_{j2},p_{2j},q_{j2},q_{2j},\widetilde{p}_{j2},\widetilde{p}_{2j},\widetilde{q}_{j2},\widetilde{q}_{2j} = 0,~~\forall j \in \{1,2,3\},\label{eq:pqc20}\\
&q_{11}\neq0,\qquad q_{33}\neq0,\label{eq:pqc21}\\
&4(p_{11}q_{13}-p_{13}q_{11})(p_{33}q_{13}-p_{13}q_{33})~+~\nonumber\\
&\qquad\qquad\qquad\qquad\qquad\qquad(p_{11}q_{33}-p_{33}q_{11})^{2}\neq 0\label{eq:pqc22}.
\end{align}
\end{lemma}

\noindent {\bf Proof of Lemma \ref{lem:PQ}:} We re-write (\ref{eq:xPx/xQx}) as
\begin{equation}
\left(\xvec_{i}^{T}P\xvec_{i}\right)\cdot\left(\xvec_{i}^{T}\widetilde{Q}\xvec_i\right)=\left(\xvec_{i}^{T}\widetilde{P}\xvec_{i}\right)\cdot\left(\xvec_{i}^{T}Q\xvec_{i}\right),
\label{eq:(xPx)(xQx)}
\end{equation}
and note that this is fourth-order polynomial equation in $\xvec_i$. Combining these equations $\forall \xvec_i \in \Omega$, we have
\begin{equation}
  \label{eq:polyform}
  V_\Omega C_{(P,Q,\widetilde{P},\widetilde{Q})} = 0,
\end{equation}
where $C_{(P,Q,\widetilde{P},\widetilde{Q})}\in\mathbb{R}^{15}$ are the coefficients of the polynomial, and are of the form $\left(p_{ij}\widetilde{q}_{kl}-\widetilde{p}_{ij}q_{kl}\right)$. Since $V_\Omega$ is rank 15, the above equation implies that $C_{(P,Q,\widetilde{P},\widetilde{Q})} = 0$. We now consider different sets of coefficients to prove the lemma.

~

\noindent {\bf Case 1.} First look at the coefficients of $x^{4},y^{4},x^{3}y,xy^{3},x^{2}y^{2}$:
\begin{align}
x^{4}:\quad & p_{11}\widetilde{q}_{11}=\widetilde{p}_{11}q_{11}\label{eq:x4}\\
y^{4}:\quad & p_{22}\widetilde{q}_{22}=\widetilde{p}_{22}q_{22}\label{eq:y4}\\
x^{3}y:\quad & p_{11}\widetilde{q}_{12}+p_{12}\widetilde{q}_{11}=\widetilde{p}_{11}q_{12}+\widetilde{p}_{12}q_{11}\label{eq:x3y}\\
xy^{3}:\quad & p_{22}\widetilde{q}_{12}+p_{12}\widetilde{q}_{22}=\widetilde{p}_{22}q_{12}+\widetilde{p}_{12}q_{22}\label{eq:xy3}\\
x^{2}y^{2}:\quad & p_{11}\widetilde{q}_{22}+4p_{12}\widetilde{q}_{12}+p_{22}\widetilde{q}_{11}=\nonumber \\
 & \qquad\widetilde{p}_{11}q_{22}+4\widetilde{p}_{12}q_{12}+\widetilde{p}_{22}q_{11}\label{eq:x2y2}
\end{align}
Since $q_{11}\neq0,q_{22}\neq0$, we can define
$t=\widetilde{q}_{11}/q_{11}$ and $s=\widetilde{q}_{22}/q_{22}$. Then
(\ref{eq:x4}) and (\ref{eq:y4}) gives us
\begin{equation}
\widetilde{q}_{11}=q_{11}t,\quad\widetilde{p}_{11}=p_{11}t,\quad\widetilde{q}_{22}=q_{22}s,\quad\widetilde{p}_{22}=p_{22}s.
\end{equation}
Substitute into (\ref{eq:x3y}), (\ref{eq:xy3}), and (\ref{eq:x2y2}),
we have
\begin{align}
-q_{11}\widetilde{p}_{12}+p_{11}\widetilde{q}_{12} & =(p_{11}q_{12}-p_{12}q_{11})t,\notag\\
(p_{12}q_{22}-p_{22}q_{12})s -q_{22}\widetilde{p}_{12} +p_{22}\widetilde{q}_{12} & =0,\notag\\
(p_{11}q_{22}-p_{22}q_{11})s -4q_{12}\widetilde{p}_{12} + 4p_{12}\widetilde{q}_{12} & =(p_{11}q_{22}-p_{22}q_{11})t.
\end{align}
This can be thought of as a linear system of equations on $\left(s,\widetilde{p}_{12},\widetilde{q}_{12}\right)$, with
one obvious solution $(t,p_{12}t,q_{12}t)$. This solution will be unique when the corresponding coefficient matrix is non-singular, \ie,
\begin{equation}
\det\left|\begin{array}{ccc}
0 & -q_{11} & p_{11}\\
(p_{12}q_{22}-p_{22}q_{12}) & -q_{22} & p_{22}\\
(p_{11}q_{22}-p_{22}q_{11}) & -4q_{12} & 4p_{12}
\end{array}\right|\neq0.
\end{equation}
Expanding this gives us (\ref{eq:pq12}), and therefore we have
\begin{equation}
[\widetilde{p}_{11},\widetilde{p}_{12},\widetilde{p}_{22}\widetilde{q}_{11},\widetilde{q}_{12},\widetilde{q}_{22}]=t[p_{11},p_{12},p_{22},q_{11},q_{12},q_{22}].
\end{equation}

This approach can be used to show the same relationship for other
terms in $P,Q,\widetilde{P},\widetilde{Q}$. Specifically, the coefficients of $\{x^{4},x^{3},x^{2},x,1\}$ give us that $[q_{11},q_{13},q_{33},p_{11},p_{13},p_{33}]$
and $[\widetilde{q}_{11},\widetilde{q}_{13},\widetilde{q}_{33},\widetilde{p}_{11},\widetilde{p}_{13},\widetilde{p}_{33}]$
are proportional, and since they are linked by $q_{11}$ and $\widetilde{q}_{11}$,
the constant of proportionality must also be $t$. Similarly, looking at the coefficients of $\{y^{4},y^{3},y^{2},y,1\}$ gives us that
$[q_{22},q_{23},q_{33},p_{33},p_{23},p_{33}]$ and $[\widetilde{q}_{22},\widetilde{q}_{23},\widetilde{q}_{33},\widetilde{p}_{33},\widetilde{p}_{23},\widetilde{p}_{33}]$
are proportional, with $q_{22}$ and $\widetilde{q}_{22}$ linking the proportionality constant to $t$.

~

\noindent {\bf Case 2.} For this case, we need to only look at the coefficients of $\{x^{4},x^{3},x^{2},x,1\}$, which gives us $[q_{11},q_{13},q_{33},p_{11},p_{13},p_{33}] = t[\widetilde{q}_{11},\widetilde{q}_{13},\widetilde{q}_{33},\widetilde{p}_{11},\widetilde{p}_{13},\widetilde{p}_{33}]$.\hfill\IEEEQEDclosed

~

\noindent {\bf Proof of Lemma \ref{lem:AllA}:} Without loss of generality, we rotate and translate the co-ordinate system so that $a_{3} = 0$, and $(0,0) \in \Omega$, and define $P=A^{T}ll^{T}A$, $Q=A^{T}A$, $\widetilde{P}=\widetilde{A}^{T}\widetilde{l}\ \widetilde{l}^{T}\widetilde{A}$
and $\widetilde{Q}=\widetilde{A}^{T}\widetilde{A}$. We consider two cases corresponding to the rank of $A$.

~

\noindent{\bf Case 1.} Rank$(A) = 3$:~ We apply case 1 of Lemma~\ref{lem:PQ} by showing that the conditions \eqref{eq:q-neq0}-\eqref{eq:pq23} hold:
\begin{enumerate}[\IEEEsetlabelwidth{~}]
\item Since $A$ is invertible, we have $q_{11}=4a_1^2 \neq 0$ and $q_{22} = 4a_2^2 \neq 0$. Also, $q_{33}=a_{4}^{2}+a_{5}^{2}+1\neq0$, and therefore, \eqref{eq:q-neq0} is satisfied.
\item For (\ref{eq:pq12}) to be satisfied, we need 
\begin{equation}
256a_{1}^{4}a_{2}^{4}\left(l_{x}^{2}+l_{y}^{2}\right)^{2}\neq0,
\end{equation}
where $l = [l_x,l_y,l_z]$. Since $A$ is invertible, $a_{1}\neq0$ and
$a_{2}\neq0$. Note that \eqref{eq:pq12} is violated if $l_{x}=l_{y}=0$ and
$\widetilde{l}_x=\widetilde{l}_y=0$ (if not the latter, we can switch $\{a,l\}$,
and $\{\widetilde{a},\widetilde{l}\}$), but in that case, it is easy to see that
\begin{equation}
A^{T}ll^{T}A = \left[\begin{array}{ccc}
0 & 0 & 0\\
0 & 0 & 0\\
0 & 0 & l_z^2
\end{array}\right],\ 
\widetilde{A}^{T}\widetilde{l}\ \widetilde{l}^{T}\widetilde{A}
 = \left[\begin{array}{ccc}
0 & 0 & 0\\
0 & 0 & 0\\
0 & 0 & \widetilde{l}_z^2
\end{array}\right],
\end{equation}
which in turn implies
$A^Tll^TA=t\widetilde{A}^{T}\widetilde{l}\ \widetilde{l}^{T}\widetilde{A}$,
with $t=l_z^2/\widetilde{l}_z^2$.
\item For (\ref{eq:pq13})-(\ref{eq:pq23}) to be satisfied, we need 
\begin{equation}
16a_{1}^{4}\left((a_{5}^{2}+1)l_{x}^{2}+(l_{z}-a_{5}l_{y})^{2}\right)^{2}\neq0,
\end{equation}
\begin{equation}
16a_{2}^{4}\left((a_{4}^{2}+1)l_{y}^{2}+(l_{z}-a_{4}l_{x})^{2}\right)^{2}\neq0.
\end{equation}
Since $a_{1},a_{2}\neq 0$, these conditions will be violated when $l_{x}=0$
and $l_{z}-a_{5}l_{y}=0$; or $l_{y}=0$
and $l_{z}-a_{4}l_{x}=0$, respectively. But these cases can be ruled out, since they result in the point $(0,0)$ being in shadow.
\end{enumerate}
Therefore, from case 1 of Lemma~\ref{lem:PQ} we have that
\begin{equation}\label{eq:AA-eq-tAA}
A^Tll^TA = t\widetilde{A}^T\widetilde{l}\widetilde{l}^T\widetilde{A},~~A^TA = t\widetilde{A}^T\widetilde{A}.
\end{equation}
To show $t=1$, we first look at the top-left $2\times2$ block of the matrix
\begin{equation}
\left[\begin{array}{cc}
4a_1^2 & \\
& 4a_2^2
\end{array}\right]=t\left[\begin{array}{cc}
4\ta_1^2+\ta_3^2 & 2\ta_3(\ta_1+\ta_2) \\
2\ta_3(\ta_1+\ta_2) & 4\ta_2^2+\ta_3^2
\end{array}\right],
\end{equation}
which implies
\begin{equation}
a_1=p_1\sqrt{t}\ta_1,\ a_2=p_2\sqrt{t}\ta_2,\ \ta_3=0,
\end{equation}
\vfill\pagebreak\noindent
with $p_1=\pm1,p_2=\pm1$. Next compare the $(1,3)$ and $(2,3)$ entry of the
matrices \eqref{eq:AA-eq-tAA}, we have
\begin{equation}
\begin{cases}
2a_1a_4=&2t\ta_1\ta_4\\
2a_2a_5=&2t\ta_2\ta_5
\end{cases},\quad\Rightarrow\quad
\begin{cases}
a_4=&p_1\sqrt{t}\ta_4\\
a_5=&p_2\sqrt{t}\ta_5
\end{cases}.
\end{equation}
Finally, look at the $(3,3)$ entry of the matrices in \eqref{eq:AA-eq-tAA}
\begin{equation}
1+a_4^2+a_5^2=t(1+\ta_4^2+\ta_5^2),\quad\Rightarrow\quad t=1.
\end{equation}

~

\noindent {\bf Case 2.} Rank$(A)=2$:~Again, without loss of generality, we assume that the rank deficiency in $A$ is caused by $a_2$ being equal to $0$. Before we can apply case 2 of Lemma~\ref{lem:PQ}, we need to show that there is no possible solution for $(\widetilde{a},\widetilde{l})$ where $\widetilde{a}_2 \neq 0$, or $\widetilde{a}_3 \neq 0$. To do so, we look at the expression for $I_\xvec$ in terms of $a$ and $l$:
\begin{equation}
I_\xvec=\frac{-(2a_{1}x+a_{4})l_{x}-a_{5}l_{y}+l_{z}}{\sqrt{(2a_{1}x+a_{4})^{2}+a_{5}^{2}+1}}.\label{eq:cxy}
\end{equation}
Note that the intensity here is independent of the coordinate $y$. Since $(\widetilde{a},\widetilde{l})$ produce the same set of intensities, they too must be independent of $y$, which implies that $\widetilde{a}_2=\widetilde{a}_3 = 0$.

We can then simply apply case 2 of Lemma~\ref{lem:PQ}, using the same approach as in case 1 above, where \eqref{eq:pqc20},\eqref{eq:pqc21} are directly satisfied by the constraints on $a$ and that $\widetilde{a}_2,\widetilde{a}_3 = 0$, and \eqref{eq:pqc22} is satisfied by $a_1\neq 0$, and the constraint that the point $(0,0)$ not be in shadow. \hfill\IEEEQEDclosed

}

\end{document}